\documentclass[runningheads]{llncs}
\usepackage[T1]{fontenc}
\usepackage{graphicx}
\usepackage{booktabs}
\usepackage{subcaption}
\usepackage[misc]{ifsym}
\usepackage{tcolorbox}

\usepackage{mwe}
\usepackage[nolist]{acronym}
\usepackage{multirow}
\usepackage{xcolor}
\usepackage{paralist}
\usepackage{amssymb}
\usepackage{adjustbox}
\usepackage{pifont}
\usepackage{cuted}
\usepackage{hyperref}
\newcommand{\cmark}{\ding{51}}%
\newcommand{\xmark}{\ding{55}}%
\newcommand{\rNet}{RescueNet}
\newcommand{\fNet}{FloodNet}
\newcommand{\gdino}{Grounding DINO\xspace}
\newcommand{\internvl}{InternVL 3.5 8B\xspace}
\newcommand{\qwenvl}{QwenVL 3 8B\xspace}
\newcommand{\gpt}{GPT-5.1\xspace}
\newcommand{\gemini}{Gemini 3 Pro\xspace}
\providecommand{\email}[1]{#1}
\acrodef{vqa}[VQA]{Visual Question Answering}
\acrodef{uav}[UAV]{Unmanned Aerial Vehicle}
\acrodef{llm}[LLM]{Large Language Model}
\acrodef{lvlm}[LVLM]{Large Vision-Language Model}
\acrodef{cv}[CV]{Computer Vision}
\acrodef{rmse}[RMSE]{Root Mean Squared Error}
\acrodef{rsqr}[$R^2$]{Coefficient of Determination}

\begin{document}

\title{Assessing the Benefits of Combining Advanced Deep Learning Techniques for Post-Disaster Building Damage Assessment from UAV Imagery}

\toctitle{Assessing the Benefits of Combining Advanced Deep Learning Techniques for Post-Disaster Building Damage Assessment from UAV Imagery}

\titlerunning{Post-Disaster Building Damage Assessment from UAV Imagery}

\author{
Huy Quang Ung\inst{1}(\Letter) \and
Guillaume Habault\inst{1} \and
Roberto Legaspi\inst{1}\and 
Hao Niu\inst{1} \and \\
Lian Cao\inst{1} \and
Masato Taya\inst{1}
}

\tocauthor{Huy~Quang~Ung, Guillaume~Habault, Roberto~Legaspi, Hao~Niu, Lian~Cao, and Masato~Taya}

\authorrunning{Ung et al.}

\institute{KDDI Research, Inc., Fujimino, Japan \\
\email{\{xhu-ung,xgu-habault\}@kddi.com},
\email{ro-legaspi@kddi-research.jp},
\email{\{ha-niu,xli-cao,ma-taya\}@kddi.com}}

\maketitle              

\begin{abstract}
Rapid and accurate post-disaster building damage assessment is essential, yet remains a challenging task.
\acf{uav} imagery offers a timely and high-resolution view of affected areas, but existing \acf{cv} models often demand large annotated datasets, generalize poorly across geographic regions and their assessment policies, and are confined to the specific tasks they were trained for.
\acfp{lvlm} offer a promising alternative through their strong reasoning and generalization capabilities, but fall short on precise, low-level perception tasks such as object detection and accurate bounding box generation.
Furthermore, they often require a substantial amount of data for effective fine-tuning on domain-specific tasks.
In this paper, we propose a hybrid framework that decouples detection from damage assessment, combining the precision of \ac{cv} models with the reasoning power of \acp{lvlm}. 
A \ac{cv} model first detects buildings and generates bounding boxes on the image that
are then passed to an \ac{lvlm} for damage classification and contextual interpretation. 
We evaluated our framework on two real-world benchmarks: \rNet{} and \fNet. 
In particular, the best combination under this framework accurately counts intact, partially damaged and completely destroyed buildings, surpassing isolated baselines by up to $2.1 R^2$ points, while requiring only limited annotated data for the detection stage.
Beyond reporting aggregate gains, we provide a detailed analysis of failure scenarios and edge cases, offering practical insights for practitioners and concrete directions for future work.
Our source code and data are publicly available to the research community via the following repository:\\\url{https://github.com/ungquanghuy-kddi/VLM_GDINO.git}

\end{abstract}

\section{Introduction}
Disaster management is increasingly critical as climate change drives the growing frequency and severity of extreme events worldwide. 
Rapid and accurate post-disaster damage assessment is essential to prioritize rescue operations, guide emergency response, and enable efficient recovery planning. 
However, conventional field surveys are time-consuming, labor-intensive, potentially unsafe, and might fail to prioritize the most critical areas due to the absence of preliminary damage identification.

Imagery captured from vehicle dash cameras, \acfp{uav}, and satellites  are valuable source of information that enables large-scale remote analysis of disaster-affected regions. 
Among these sources, \acp{uav}, which can easily navigate within disaster areas, are particularly advantageous as they offer high-resolution imagery, flexible and fast deployment: making them well-suited for rapid post-disaster reconnaissance. 
Coupled with \acf{cv} models, which have demonstrated strong performance in object detection, \ac{uav}-based imagery presents a promising pathway toward automated, timely, and cost-effective preliminary damage assessments to support disaster management efforts.

Despite this promise, several challenges remain. 
\ac{uav} imagery is inherently complex, often spanning large areas with significant variation in object scale, shape, and appearance, as well as partial occlusion by vegetation.
These factors make accurate detection and classification of building damage particularly difficult.
Beyond perception, damage level definitions vary across regions and applications (from rescue dispatch to insurance assessment) limiting the generalizability of automated systems.
While recent extended \ac{cv} models such as \gdino~\cite{liu2024grounding} enable open-set object detection in both zero-shot or few shot settings, accurately assessing building damage often requires broader contextual understanding of area-wide conditions, which purely perception-based models may struggle to capture.

Lately, \acfp{lvlm} have demonstrated remarkable reasoning and generalization capabilities with high-level tasks, including mathematical reasoning~\cite{ahn2024large}, \ac{vqa}~\cite{liu2023visual}, and common sense reasoning~\cite{suzgun2023challenging}.
Capabilities that are highly relevant to post-disaster building damage assessment task.
Furthermore, their language-guided nature allows \acp{lvlm} more flexibility to adapt more easily to varying assessment policies and damage definitions.
However, as demonstrated in~\cite{sapkota2026object,zhan2024griffon}, they often fall short on precise, low-level perception tasks such as object detection and accurate bounding box positioning, which are critical for identifying individual buildings in \ac{uav} imagery.
Besides, fine-tuning \acp{lvlm} to address this gap is impractical, requiring substantial computational resources and large annotated datasets.

To address these limitations, we propose a hybrid framework that leverages the complementary strengths of both \ac{cv} models and \acp{lvlm} for post-disaster building damage assessment from \ac{uav} imagery. 
This proposal decouples assessment into two stages: a \ac{cv}-based detection module that identifies buildings and generates precise bounding boxes, and an \ac{lvlm}-based reasoning module that classifies damage level of detected buildings as well as provides contextual interpretation. 
These high-level tasks span from identifying priority rescue zones, to generating descriptive damage analyses, or coordinating efficient repair of disaster areas.
This design preserves the spatial accuracy and efficiency of traditional \ac{cv} models while leveraging the semantic understanding and adaptability of \ac{lvlm}, without requiring full fine-tuning of either component. 
Crucially, by requiring only limited annotated data for the detection component, our framework is practical for real-world deployment where labeled data is scarce.

\textbf{Our contributions} are as follows:

\begin{itemize}
    \item We introduce a hybrid \ac{cv}-\ac{lvlm} framework that decouples building detection from damage assessment for post-disaster building damage assessment from \ac{uav} imagery. 
    \item We evaluate this framework by combining \gdino with two open-source and two proprietary \acp{lvlm}. The associated experiments target the identification and counting of intact, damaged or completely destroyed buildings on two real-world datasets. 
    \item We demonstrate that our framework outperforms isolated components, particularly under limited fine-tuning data for \gdino, highlighting the value of combining complementary model strengths.
    \item We provide a detailed analysis of our results covering failures and edge cases, offering actionable insights for practitioners and concrete directions for future research.
\end{itemize}

\section{Related Work}

\subsubsection{Computer Vision-based Damage Assessment} 
\acf{cv} models are widely used for building detection and damage assessment in both aerial and satellite imagery. 
Popular detection models such as the YOLO series~\cite{reis2023real} and Mask R-CNN~\cite{he2017mask} have been applied to instance segmentation for disaster response~\cite{sirma2025drespnet} and fine-grained object-level damage assessment~\cite{sadiq2023towards}.
In addition, segmentation models such as U-Net~\cite{ronneberger2015u} have been employed to detect damaged buildings by comparing pre- and post-disaster satellite imagery~\cite{deng2022post}.

More recently, foundation open-set object detection models such as \gdino~\cite{liu2024grounding} and YOLOE~\cite{wang2025yoloe} have emerged, enabling flexible, text-guided detection without requiring class-specific training.
This capability offers strong potential for disaster response scenarios, where damage-related targets, new or unexpected object types may need to be identified.

\subsubsection{Pre-trained Large Vision-Language Models}
\acfp{lvlm} have significantly advanced multimodal understanding and reasoning across a wide range of tasks. 
Proprietary models such as OpenAI's GPT~\cite{chatgpt} or Google's Gemini~\cite{gemini3}, along with open-source models including InternVL~\cite{wang2025internvl3_5}, QwenVL~\cite{qwen3technicalreport}, and LLaVA~\cite{li2024llava}, have demonstrated strong capabilities in visual question answering, reasoning, and instruction following.
However, these general-purpose \ac{lvlm} might be primarily trained on natural images and web-scale multimodal data, which limits their effectiveness when applied directly to specialized domains such as \ac{uav} imagery and damage assessment.

Recent works adapt \acp{lvlm} to remote sensing imagery, which presents distinct challenges such as large spatial coverage and domain-specific semantics. 
GeoChat~\cite{kuckreja2024geochat} introduces a conversational framework that integrates geospatial visual encoders with \acp{llm} for interactive understanding of satellite images. 
EarthDial~\cite{soni2025earthdial} enables multi-turn dialogue about Earth observation data through instruction tuning, while TEOChat~\cite{irvinteochat} focuses on multimodal reasoning over satellite imagery using domain-specific instruction datasets. 
Although these approaches demonstrate the potential of \acp{lvlm} for remote sensing, they are designed for general Earth observation understanding and are not specifically tailored for post-disaster damage assessment tasks.
To foster the development of \acp{lvlm} for disaster response, large scale datasets have been introduced such as DisasterM3 benchmark~\cite{wangdisasterm3} and Incidents1M~\cite{weber2022incidents1m}.

\subsubsection{Hybrid Framework Combining \acp{lvlm} and \ac{cv} models} 
Hybrid framework has recently emerged as a promising direction across several \ac{cv} domains.
While \acp{lvlm} offer strong semantic reasoning and flexibility, they struggle with precise spatial localization and can be computationally demanding~\cite{sapkota2026object,zhan2024griffon}.
Conversely, dedicated \ac{cv} detectors excel at localizing objects accurately but lack the contextual understanding needed for complex classification tasks.
To address these complementary limitations, recent works have proposed hybrid frameworks that decouple spatial detection from semantic reasoning, assigning each task to the component best suited for it.
Works such as DetGPT~\cite{pi2023detgpt} and ContextDET~\cite{zang2025contextual} follow this principle by using \ac{cv} detectors to localize objects and \acp{lvlm} to interpret their context or condition.
A similar decoupling has been adopted in robotics, where frameworks such as SayCan~\cite{ahn2022can} which separate high-level task planning, handled by language models, from low-level spatial perception, handled by vision modules.
Our work extends this paradigm to the remote sensing domain, decoupling building detection (handled by a \ac{cv} model) from the contextually demanding task of post-disaster damage assessment (handled by a \ac{lvlm}).
To the best of our knowledge, this is the first application of such a hybrid framework to \ac{uav}-based building damage assessment.

\section{Methodology: Combining VLM and CV}
\begin{figure}[t]
    \centering
    \includegraphics[width=0.9\linewidth]{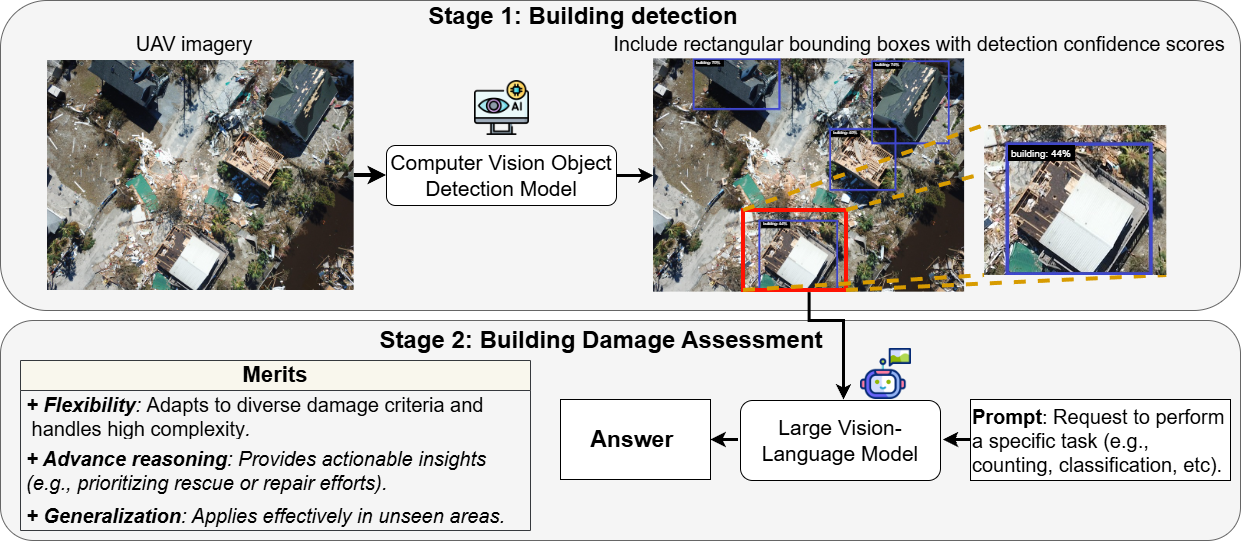}
    \caption{Overview of our proposed hybrid framework along with its key merits: \emph{flexibility}, \emph{advanced reasoning}, and \emph{generalization}. Stage 1 uses a \ac{cv} model to detect buildings and generate bounding boxes with confidence scores. Stage 2 uses an \ac{lvlm}, guided by a task-specific prompt, to perform damage assessment.}\label{fig:framework}
\end{figure}

This section presents the rationale behind our hybrid framework, details its design, and describes the experimental setup used to evaluate it.

\subsection{Rationale}
\ac{cv} models and \acp{lvlm} each excel at fundamentally different aspects of the building damage assessment pipeline, making neither sufficient on its own.
\ac{cv} models achieve strong performance in spatial localization tasks: detecting objects of varying classes and generating accurate bounding boxes.
However, they struggle when classification requires higher-level contextual understanding beyond object-level cues.
In contrast, \acp{lvlm} are primarily designed for image understanding and reasoning rather than precise spatial detection~\cite{sapkota2026object,zhan2024griffon}. 
Yet they offer significant advantages for difficult assessment cases through semantic reasoning and contextual interpretation of scene-wide conditions. 
As a consequence, neither approach alone is sufficient to fully address the task.
Our framework exploits their complementary strengths into a unified framework: a \ac{cv} model handles building detection, while an \ac{lvlm} performs damage-level assessment.
Each component operates within its domain of competence.

\subsection{Framework Design}
To enable robust post-disaster building damage assessment from \ac{uav} imagery, we propose a two-stage pipeline that integrates a \ac{cv} detector with an \ac{lvlm} reasoner, as illustrated in Figure~\ref{fig:framework}.

\textbf{Stage 1: Building Detection.} 
Given an input \ac{uav} image, a \ac{cv}-based object detector identifies buildings in the scene, enclosing each detected instance in a rectangular bounding box along an confidence scores.
This stage leverages the strong spatial localization capability of modern \ac{cv} detectors to reliably identify buildings even in cluttered, disaster-affected environments~\cite{sadiq2023towards}.
The detector can be fine-tuned on domain-specific data to further improve detection accuracy.
The image, overlaid with bounding boxes, is then passed to the second stage.

\textbf{Stage 2: Damage Assessment.}
The image enhanced with bounding boxes and confidence scores alongside a task-specific prompt are passed as input to the \ac{lvlm}, which performs per-building damage classification through multi-modal reasoning.
By receiving explicit spatial proposals from Stage 1, the \ac{lvlm} can focus its reasoning on individual building regions as well as overall contextual information.

\textbf{Interpretation.}
Building-level assessments produced by the \ac{lvlm} are aggregated in order to generate statistics of affected and unaffected buildings in the target image. 
These outputs then support a range of high-level tasks: estimating the overall severity of damage across an area, prioritizing regions for urgent rescue, generating actionable situation reports, and planning how to restore critical services (e.g., electricity, water).
The language-guided nature of the \ac{lvlm} makes such outputs readily adaptable to different assessment policies and operational contexts

\subsection{Experimental Design}
\label{sec:exp_design}

\textbf{Model Selection.}
Training \ac{cv} models typically require large amount of annotated data in order to achieve strong task-specific performance.
In contrast, a foundational model like \gdino, which have already been pretrained, can be effectively adapted with limited labeled data.
Liu et al.~\cite{liu2024grounding} demonstrated that fine-tuned it outperforms Mask R-CNN (pure \ac{cv} model) under limited amount of annotated data.
Such a finding is directly relevant to post-disaster scenarios where annotated data is scarce.
Therefore, in our experiments, we adopt \gdino as the detection component of our framework.
We hypothesize that fine-tuning it for building detection in \ac{uav} imagery will outperform training a \ac{cv} model from scratch, especially with limited data.

\textbf{Task and Metrics.}
Our experiments target the counting of buildings across all categories (affected and unaffected by the disaster).
For each image, per-class outputs are combined in order to compute different metrics.
Performance on the counting task is evaluated using \ac{rmse} and \ac{rsqr}.
To provide additional perspective on such performances, we report Average Precision (AP) at Intersection over Union (IoU) larger than $0.75$ (denoted as $AP^{75}$) to evaluate the spatial accuracy of the building detection task.

\textbf{Models Evaluated.}
In this study, we combine \gdino with four \acp{lvlm}: two proprietary (\gpt, \gemini) and two open-source (\qwenvl, \internvl).
Lightweight variants of the open-source models are deliberately chosen to assess the feasibility of edge deployment, such as integration directly onto \ac{uav} platforms.

\textbf{Baselines.}
We compare our framework against the following baselines:
\begin{itemize}
    \item \textit{Random}: randomly guesses the number of buildings for each damage level within the observed range. It serves as a lower-bound reference.
    \item \textit{Random\_[class]}: simulates single-class damage detection model, designed to expose models that, under class imbalance, focus on the majority class.
    \item G-DINO* (\gdino alone): \gdino fine-tuned to perform both detection and damage assessment of all classes, followed by an algorithm counting the number of detected bounding boxes per class. This isolates the contribution of the \ac{lvlm} reasoning stage.
    \item \ac{lvlm}-alone: the \ac{lvlm} performs detection, damage classification, and per-class counting of detected buildings, without spatial identification from a \ac{cv} detector. This isolates the contribution of the \ac{cv} detection stage.
\end{itemize}

Together, these baselines allow us to disentangle the contributions of each component and rigorously assess the value of combining them.

\textbf{Fine-tuning Subset.}
To validate our hypothesis that \gdino can achieve satisfactory performance with limited amount of labeled data for the building damage assessment task, we define two fine-tuning subsets: ``FT-100'' and ``FT-Full''.
FT-100 represents a low-data condition, comprising $100$ samples drawn from the training set ($80$\%) and the validation set ($20$\%).
These samples were randomly selected while preserving the class distribution of the full dataset as closely as possible.
FT-Full uses all available training and validation samples.

To further investigate the effect of fine-tuning subset size on performance, we conduct an ablation study in which we report results across subsets of increasing size, generated using the same sampling strategy as FT-100.
It is important to note that each subset is sampled independently: larger subsets do not necessarily contain all samples from smaller ones.
This ablation provides a fine-grained view of how the performance of our framework scales with data availability.

\textbf{\ac{lvlm} Fine-tuning.}
We deliberately keep all \acp{lvlm} in their original pretrained form without any fine-tuning.
The available training data is relatively limited, making fine-tuning prone to both overfitting and catastrophic forgetting, causing the model to potentially lose both its broad pretrained knowledge and its visual-language alignment~\cite{lin2024lora}.
Moreover, Gulati et al.~\cite{gulati2026narrow} demonstrated that parameter-efficient methods such as LoRA~\cite{hu2022lowrank} may also introduce emergent multimodal misalignment under low-data conditions, potentially degrading multimodal reasoning performance. 
Given these risks, and as our main goal is to evaluate the effectiveness of the proposed framework rather than to adapt individual \acp{lvlm}, we rely on their pretrained capabilities throughout this study.

\section{Dataset}

\begin{table}[b]
    \small
    \centering
    \resizebox{\columnwidth}{!}{%
    \begin{tabular}{c|c|c|c|ccc|c|ccccccccc}
    \multicolumn{8}{c}{\textbf{}} & \multicolumn{9}{|c}{\textbf{Class}} \\
    {} & \textbf{Disaster} & \textbf{Geographic} & \textbf{Resolution} & \multicolumn{3}{|c|}{\textbf{Split Size}} & \textbf{Annotation} & \vbox{\hbox{\multirow{2}{*}{\rotatebox[origin=c]{90}{\tiny{\textbf{Background}}}}}\vspace{18pt}}& \multirow{2}{*}{\rotatebox[origin=c]{90}{\tiny{\textbf{Pool}}}}& \multirow{2}{*}{\rotatebox[origin=c]{90}{\tiny{\textbf{Water}}}}& \multirow{2}{*}{\rotatebox[origin=c]{90}{\tiny{\textbf{Road}}}}& \vbox{\hbox{\multirow{2}{*}{\rotatebox[origin=c]{90}{\tiny{\textbf{Vehicle}}}}}\vspace{4pt}} & \multirow{2}{*}{\rotatebox[origin=c]{90}{\tiny{\textbf{Tree}}}}& \multirow{2}{*}{\rotatebox[origin=c]{90}{\tiny{\textbf{Sand}}}}& \multirow{2}{*}{\rotatebox[origin=c]{90}{\tiny{\textbf{Grass}}}}& \vbox{\hbox{\multirow{2}{*}{\rotatebox[origin=c]{90}{\tiny{\textbf{Building}}}}}\vspace{8pt}}\\
    \textbf{Dataset} & \textbf{Type} & \textbf{Scope} & \textbf{(pixel)} & \textbf{Tr} & \textbf{Val} & \textbf{Te} & \textbf{Methodology} & {} & {}& {}& {}& {}& {}& {}& {}& {}\\\hline
    {\rNet} & {Hurricane} & {United States} & \multirow{2}{*}{3000x4000} & {3595} & {449} & {450} & {pixel-level} & {\cmark} & {\cmark} & {\cmark} & {\cmark} & {\cmark} & {\cmark} & {\cmark} & {\textcolor{lightgray}{\xmark}} & \textcolor{green}{\cmark} \\\cline{1-3}\cline{5-7}\cline{9-17}
    {\fNet} & {Flood} & {Texas/Louisiana} & {} & {1445} & {450} & {448} & {labeling} & {\cmark} & {\cmark} & {\cmark} & {\cmark} & {\cmark} & {\cmark} & {\textcolor{lightgray}{\xmark}} & {\cmark} & {\textcolor{green}{\cmark}} \\\hline
    \end{tabular}
    }
    \caption{Comparison of \rNet{} and \fNet{} dataset characteristics, including disaster type, image resolution, number of samples per data splits, and semantic class coverage.}
    \label{tab:dataset_info}
\end{table}

We evaluate our framework on two high-resolution post-disaster \ac{uav} imagery datasets: \rNet{} and \fNet.
Both were collected in the aftermath of hurricanes in the United States and provide fine-grained pixel-level annotations for semantic segmentation, covering classes representative of post-disaster environments such as water, roads, vehicles, vegetation, and buildings.
Despite their shared structure, the two datasets differ in their geographic contexts, disaster type, and associated challenges, making their combination valuable for assessing the generalizability of our framework.
Table~\ref{tab:dataset_info} summarizes key information for both datasets.
A key limitation of both datasets is their geographic and event-specific scope.
As both were collected exclusively in the United States following a single event, models trained on them may not generalize well to other terrains or geographic contexts.

As this study focuses exclusively on building damage assessment, all non-building classes are discarded.
As a result, \rNet{} yields four building-related classes and \fNet{} two.

\subsection{\rNet}
\rNet~\cite{rahnemoonfar2023rescuenet} was acquired following Hurricane Michael ($2018$) and defines building damage in accordance with FEMA~\cite{fema} guidelines, distinguishing four levels: \emph{no damage}, \emph{minor damage}, \emph{major damage}, and \emph{total destruction}. 
This hierarchical taxonomy makes it particularly well-suited for structural damage assessment tasks, where differentiating degrees of destruction is critical. 

\begin{figure}[h]
    \centering
    \includegraphics[width=\linewidth]{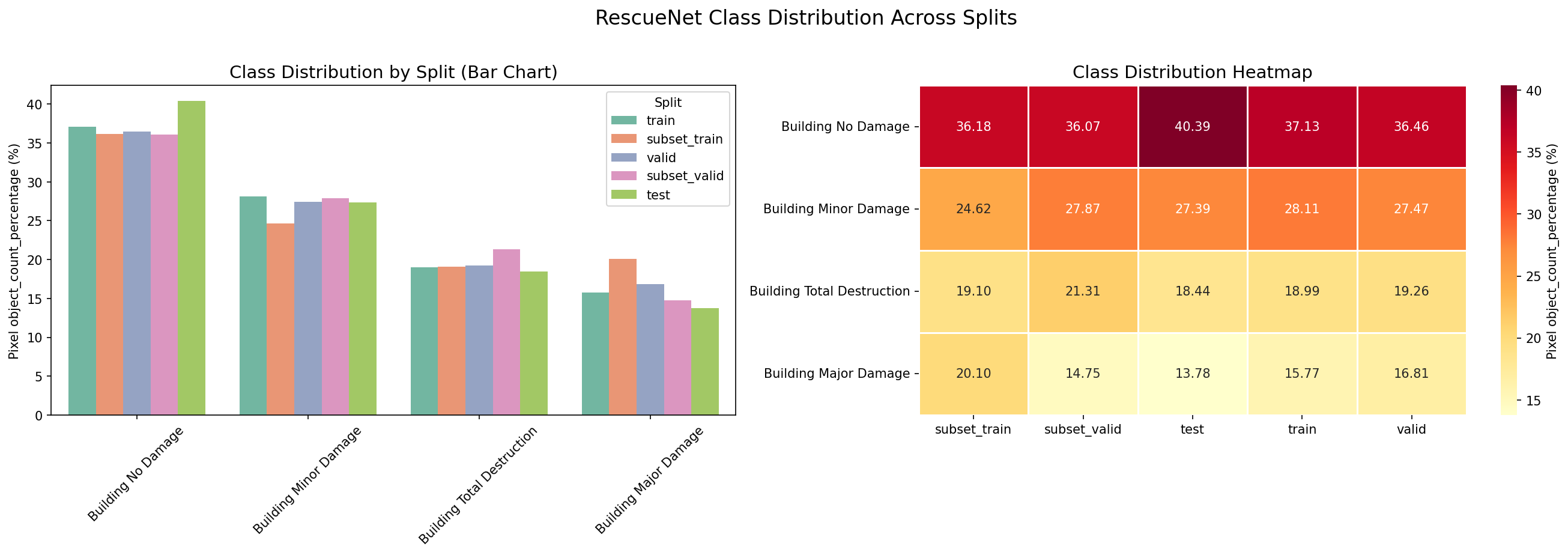}
    \caption{Distribution of average building count per image across \rNet{} data splits and the FT-100 fine-tuning subset (subset\_train, subset\_valid), shown as class-wise percentages. The FT-100 sampling strategy successfully preserves the original class distribution, with the \emph{no damage} class remaining dominant across all splits.}
    \label{fig:rescuenet_limited_distri}
\end{figure}

\textbf{Preprocessing.}
Distinguishing minor from major damage is inherently difficult from aerial imagery alone~\cite{gebre2026multimodalattentionautomateddisaster}.
Therefore, we merge these two categories into a single ``damaged'' class, reducing the classification to three levels: \emph{no damage}, \emph{damaged}, and \emph{total destruction}.
Figure~\ref{fig:rescuenet_limited_distri} illustrates the class distribution of \rNet{} and the associated FT-100 fine-tuning subset, showing that FT-100 successfully preserves the original distribution.
In addition, we observe that after merging minor and major classes, the resulting distribution is relatively balanced in terms of average building count per image.

\textbf{Challenge.}
A key difficulty in \rNet{} is the visual ambiguity between debris from totally destroyed buildings and debris transported by the hurricane.
Such a distinction is particularly hard to make without pre-disaster reference imagery.

\subsection{\fNet}
\fNet~\cite{rahnemoonfar2021floodnet} was acquired in the aftermath of Hurricane Harvey ($2017$) and defines both road and building damage along a binary dimension: \emph{flooded} and \emph{non-flooded}. 

\begin{figure}
    \centering
    \includegraphics[width=\linewidth]{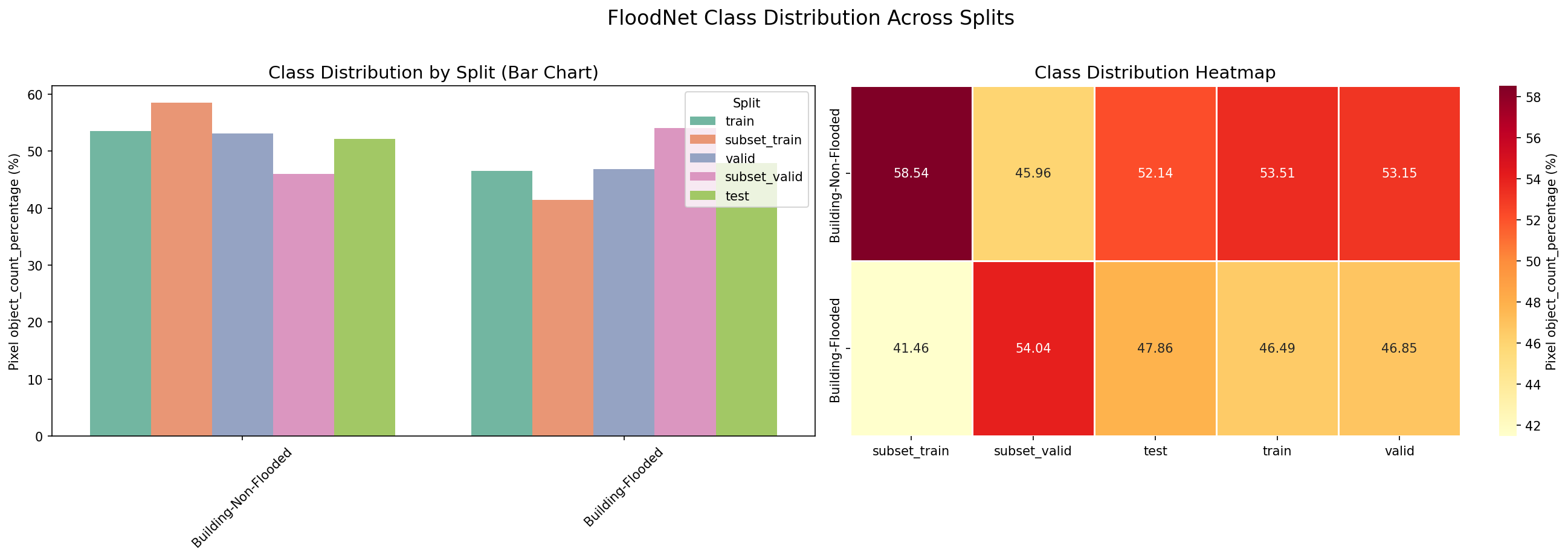}
    \caption{Distribution of average building count per image across \fNet{} data splits and the FT-100 fine-tuning subset (subset\_train, subset\_valid), shown as class-wise percentages. Classes are relatively balanced in terms of average building count per image and the FT-100 sampling strategy successfully preserves the original class distribution.}
    \label{fig:floodnet_limited_distri}
\end{figure}

\textbf{Preprocessing.}
In this study, we adopt the definition of flooded buildings from the original paper.
Figure~\ref{fig:floodnet_limited_distri} illustrates the class distribution of \fNet{} as well as its associated FT-100 fine-tuning subset, showing that FT-100 closely preserves the original distribution in terms of average building count per image.
This figure shows that the class distribution of  with a relatively balanced original distribution.
However, this per-object balance masks a significant image-level imbalance with over $90$\% of images containing no flooded buildings, which can bias model training toward the dominant non-flooded class.

\textbf{Challenges.}
\fNet{} presents several inherent difficulties.
Visually, distinguishing floodwater from natural water bodies can sometimes be challenging, as both appear similar in aerial imagery.
Furthermore, the presence of dense canopy cover, may obscure floodwater making assessment more difficult.
Additionally, the lack of detailed information on the annotation process limits the ability to assess annotation reliability.

\section{Results and Analysis}

This section presents our experimental settings, main results, and an extensive ablation study.
In addition, we report the standalone detection performance of \acp{lvlm} to motivate the need for a dedicated \ac{cv} module within our framework.

\subsection{Experimental Settings}
We adopt the pre-trained \gdino implementation from~\cite{mmdetection}, configured with a Swin-T backbone.
For simplicity, we refer to \gdino as G-DINO in the rest of the paper.
Fine-tuning hyperparameters follow the default settings in~\cite{mmdetection}.
At inference, only bounding boxes with a confidence score above $0.3$ are retained.

\begin{table*}[t]
    \centering
    \resizebox{0.7\columnwidth}{!}{%
    \begin{tabular}{c|l|l}
    \toprule
         \textbf{Task} &  \textbf{\rNet} & \textbf{\fNet} \\
    \midrule
         Counting: &  (1) Intact building& (1) Non-flooded building\\
         
         &  (2) Partially damaged building& (2) Flooded building\\
    
 & (3) Totally destroyed building&\\
    \bottomrule
    \end{tabular}
    }
    \caption{List of our objectives for the counting task.}
    \label{tab:task_building}
\end{table*}

Table~\ref{tab:task_building} presents the building damage classes considered in the counting task. 
For each \ac{lvlm} query, we simply prompt the model to count buildings of a single damage class per conversation turn.
The final results are computed by aggregating the per-class responses across all conversation turns. 

\subsection{\ac{lvlm} performance on Building Detection} 
Table~\ref{tab:lvlm_det} reports zero-shot detection performance on the \rNet~and \fNet~test sets for \qwenvl, \internvl, \gpt, \gemini, and G-DINO.
With the exception of \gemini, which achieves an $AP^{75}$ of $0.36$ on \rNet, all other \acp{lvlm} have their scores close to zero.
All these scores are significantly worse than G-DINO, 
\begin{inparaenum}[(1)]
    \item confirming that object detection remains a key weakness of general-purpose \acp{lvlm}; and 
    \item providing empirical support for our framework's core design choice: decoupling detection from reasoning by delegating detection to a dedicated \ac{cv} model rather than relying on \acp{lvlm}.
\end{inparaenum}

\begin{table}
    \centering
    \small
    \resizebox{0.9\columnwidth}{!}{%
        \begin{tabular}{l|r|r|r|r||r}
            \toprule
            \textbf{Dataset} & \textbf{\qwenvl} & \textbf{\internvl} & \textbf{\gpt} & \textbf{\gemini} & \textbf{G-DINO} \\
            \midrule
            \textbf{\rNet} & {0.0003} & {0.0002} & {0.0003} & {0.3625} & \textbf{0.4991} \\
            \midrule
            \textbf{\fNet} & {0.0003} & {0.0001} & {0.0002} & {0.0731} & \textbf{0.4453} \\
            \bottomrule
        \end{tabular}
    }
    \caption{Zero-shot building detection performance (${AP}^{75}$) of \acp{lvlm} and a dedicated \ac{cv} model (G-DINO) on the \rNet~and \fNet~test sets.}
    \label{tab:lvlm_det}
\end{table}

\begin{table*}[t]
\centering
\small
\resizebox{0.9\columnwidth}{!}{%
\begin{tabular}{l|c|rrr||rrr}
\toprule
\multirow{2}{*}{\textbf{Method}}  &\multirow{2}{*}{\textbf{Fine-tuning}}& 
\multicolumn{3}{c||}{\textbf{\rNet}}& 
\multicolumn{3}{c}{\textbf{\fNet}}\\&& RMSE ($\downarrow$) & R$^2$ ($\uparrow$) 
  &$AP^{75}$($\uparrow$)& RMSE ($\downarrow$) & R$^2$ ($\uparrow$) 
  &$AP^{75}$($\uparrow$)\\
\midrule
\textit{Random} &\textcolor{lightgray}{\xmark} 
& 7.044&  -23.760&--& 24.117&  -39.036&--
\\

\textit{Random\_intact} &\textcolor{lightgray}{\xmark} 
& 4.221&  -7.888&--& --&  --&--
\\

\textit{Random\_damaged} &\textcolor{lightgray}{\xmark} 
& 4.268&  -8.089&--& --&  &--
\\

\textit{Random\_destroyed} &\textcolor{lightgray}{\xmark}  
& 4.478&  -9.007&--& --&  --&--
\\
 \textit{Random\_non\_flooded} &\textcolor{lightgray}{\xmark} 
& --& --& --& 16.885& -18.624&--
\\
 \textit{Random\_flooded} &\textcolor{lightgray}{\xmark} 
& --& --& --& 17.696& -20.556&--\\
\midrule
\multirow{3}{*}{G-DINO*} &{Not FT}& 2.955&  -3.356
&0.118
& 3.148&  0.318
&0.118
\\
  &{FT-100}& 2.315&  -1.673
&0.420
& 3.768& 0.023
&0.420
\\
  &{FT-Full}& 0.973&  0.528&0.677
& 2.041& 0.713&0.677
\\
\midrule
 \qwenvl &\textcolor{lightgray}{\xmark}& 2.759&  -2.798
&--& 2.613& 0.529
&--
\\
 \internvl &\textcolor{lightgray}{\xmark}& 2.215&  -1.447
&--& 1.871& 0.758
&--
\\
 \gpt &\textcolor{lightgray}{\xmark}& 1.328& 0.120& --& 2.611& 0.531
&--\\
 \gemini &\textcolor{lightgray}{\xmark}& 2.106&  -1.213&--& 3.790& 0.011
&--\\ \midrule
 \multirow{3}{*}{\qwenvl + G-DINO}& {Not FT}& 2.957& -3.363& 0.499& 3.078& 0.348&0.445\\
 
 &{FT-100} 
& 2.216&  -1.450
&0.710& 2.781& 0.467
&0.772\\
  &{FT-Full} 
& 2.181& -1.374& 0.839
& 2.832& 0.448
&0.840
\\ \midrule
 \qwenvl + \textit{GTBBox} &\textcolor{lightgray}{\xmark} 
& 1.967& -0.930& 1.000& 2.901& 0.421&1.000\\ \midrule
 \multirow{3}{*}{\internvl+ G-DINO}& {Not FT}& 2.152& -1.310& 0.499& 3.249& 0.273&0.445\\
  
 &{FT-100} 
& 1.993&  -0.981
&0.710& 2.954& 0.399
&0.772\\
  &{FT-Full} 
& 2.016& -1.028& 0.839
& 2.912& 0.416
&0.840
\\ \midrule
 \internvl + \textit{GTBBox} &\textcolor{lightgray}{\xmark} 
& 1.878& -0.759& 1.000& 2.766& 0.473&1.000\\  \midrule
 \multirow{3}{*}{\gpt+ G-DINO}& {Not FT}& 1.349& 0.092& 0.499& 2.481& 0.576&0.445\\
  
 &{FT-100} 
& 1.181& 0.304
& 0.710& 2.538& 0.557
&0.772\\
  &{FT-Full} 
& 1.219& 0.258& 0.839
& 2.700& 0.498
&0.840
\\ \midrule
 \gpt + \textit{GTBBox} &\textcolor{lightgray}{\xmark} 
& 1.157& 0.333& 1.000& 2.453& 0.586
&1.000\\ \midrule
 \multirow{3}{*}{\gemini+ G-DINO} & {Not FT}     
& 1.132& 0.361& 0.499& 2.889& 0.425&0.445\\
  
 &{FT-100} & 0.990&  0.511&0.710& 2.645& 0.518
&0.772\\
  &{FT-Full} 
& 1.105& 0.391& 0.839
& 2.413& 0.599
&0.840
\\ \midrule
 \gemini + \textit{GTBBox} &\textcolor{lightgray}{\xmark} & 0.918& 0.579& 1.000& 2.838& 0.445
&1.000\\
\bottomrule
\end{tabular}
}

\caption{Main results for \ac{lvlm}-alone and \ac{lvlm}+G-DINO configurations on the \rNet~and \fNet~test set. ``\textit{GTBBox}'' represents perfect bounding boxes extracted from ground-truth. ``Not FT'', ``FT-100'' and ``FT-Full'' denote G-DINO (or G-DINO*) without fine-tuning, fine-tuned with 100 samples, and fine-tuned with the full dataset, respectively.}
\label{tab:all_result}
\end{table*}

\subsection{Main Results}

\subsubsection{\rNet}

Table~\ref{tab:all_result} presents the performance of the proposed framework (using G-DINO as detector) across all \acp{lvlm} and fine-tuning configurations.
To establish an upper bound, we additionally report results when ground-truth bounding boxes (GTBBox) are provided directly to the \ac{lvlm} component, simulating a perfect detector.

Several clear trends emerge. 
First, naive baselines perform poorly across all metrics, confirming that simple guessing strategies or single-class focus cannot capture the distribution of building damage levels.
Second, G-DINO* (designed to perform both detection and damage classification jointly) improves with more fine-tuning data but consistently underperforms the proposed hybrid framework.
Third, \ac{lvlm}-alone baselines struggle to reliably detect and count buildings directly, yielding unstable performance across models compare to the proposal.
These observations demonstrate that decoupling localization from reasoning is beneficial.

The proposed \ac{lvlm}+G-DINO framework consistently outperforms all baselines, already achieving strong performance with limited fine-tuning data.
Notably, our framework achieves the largest gains using \gemini and fine-tuning G-DINO with FT-100 -- up to $1.6$ and $2.1$ $R^2$ points over the \gemini alone and G-DINO*, respectively.
Such results demonstrate the effectiveness of the hybrid approach even under low-data conditions.
As G-DINO is progressively fine-tuned, bounding box quality improves ($AP^{75}$ approaching $1$), which further benefits \ac{lvlm} reasoning and counting, narrowing the gap toward the GTBBox upper bound.

\subsubsection{FloodNet} 
Table~\ref{tab:all_result} reveals a similar overall trend for proprietary \acp{lvlm}, which benefit from integration with G-DINO.
However, open-source \acp{lvlm} exhibit a notably different behavior. 
First, combining open-source \acp{lvlm} with G-DINO occasionally degrades performance relative to the \ac{lvlm}-alone baseline, suggesting that additional spatial context provided by bounding boxes alter models capabilities.
Second, providing GTBBox yields only marginal improvements, and in some cases even slight degradation, as observed with \gemini.
This behavior contrast sharply with \rNet, where perfect localization led to significant gains. 
This discrepancy suggests that the primary bottleneck on \fNet{} lies not in localization quality but on other visual phenomenon.
We further analyze and discuss this point in the following section.

\subsection{Ablation Study}
\subsubsection{Impact of Fine-tuning Set Size}
We vary the fine-tuning set size for G-DINO and G-DINO* across $[0, 100, 500, 1000, 2000, 4044]$ samples for \rNet{} and $[0,100,500,1000,1895]$ samples for \fNet.
Each subset is generated using the sampling strategy described in Section~\ref{sec:exp_design}, i.e., randomly selecting samples while approximately preserving the full dataset's class distribution.

Table~\ref{tab:mAP} reports the detection performance of G-DINO and G-DINO* across fine-tuning set sizes with both datasets.
Fine-tuning with as few as $100$ samples already yields a substantial improvement over the zero-shot setting, and performance continues to improve steadily as more data is added.
These results confirm that G-DINO adapts effectively to \ac{uav}-based building detection even under limited annotation budgets.

\begin{figure}[t]
    \begin{subfigure}{.45\linewidth}
    \centering
    \includegraphics[width=1.0\linewidth]{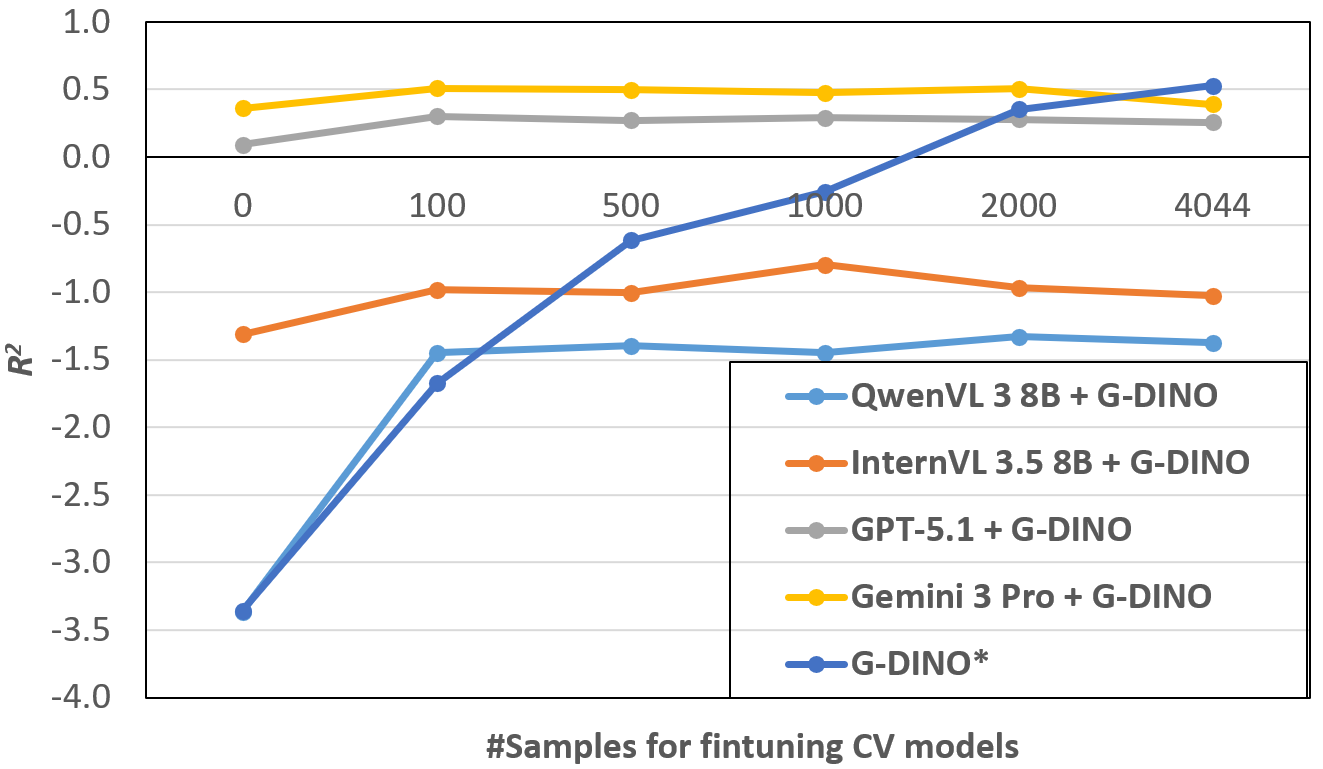}
    \caption{RescueNet test set.}
    \label{fig:sample_varis_rescuenet}
    \end{subfigure}
    \hspace{\fill}
    \begin{subfigure}{.45\linewidth}
    \centering
    \includegraphics[width=1.0\linewidth]{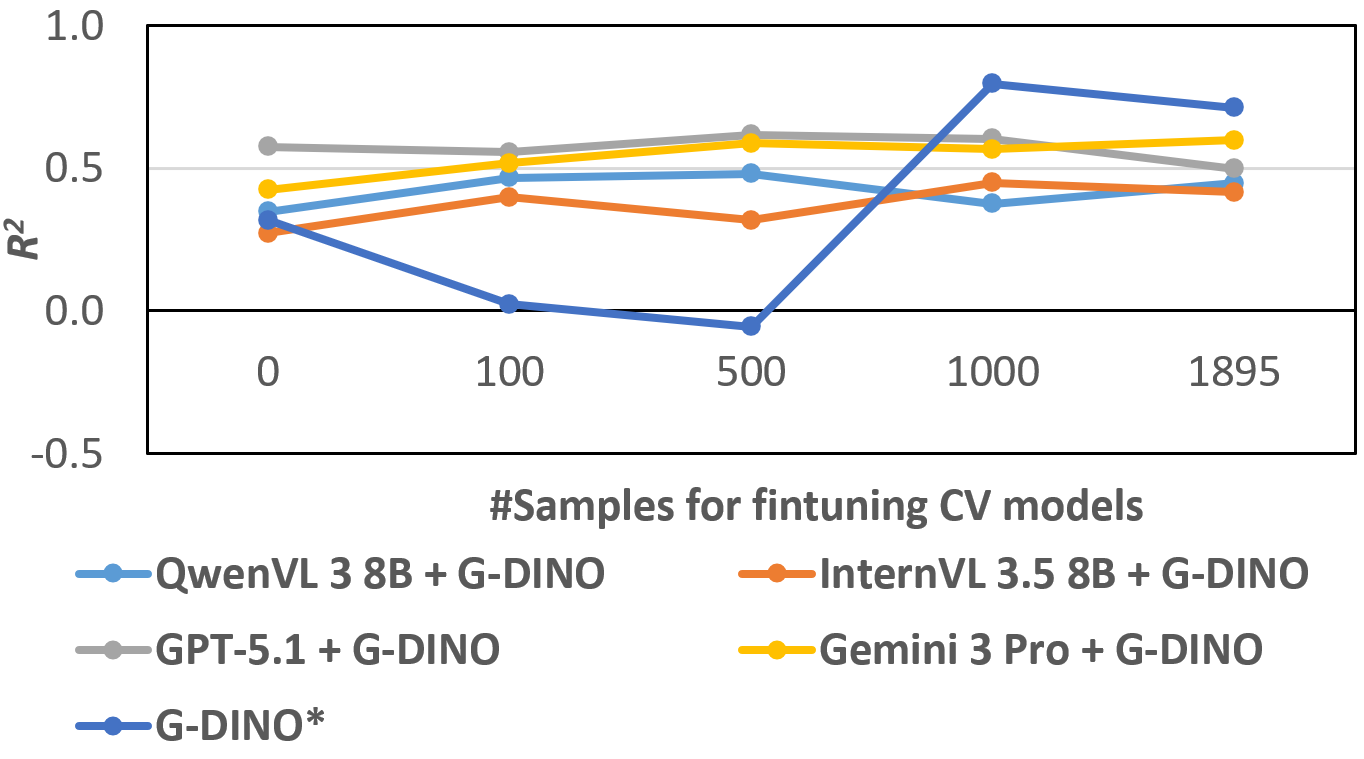}
    \caption{FloodNet test set.}
    \label{fig:sample_varis_floodnet}
    \end{subfigure}
    \caption{Evolution of $R^2$ scores (y-axis) across varying fine-tuning set sizes (x-axis) of G-DINO* and our proposed \ac{lvlm}+G-DINO framework on (a) \rNet~and (b) \fNet.}
\end{figure}

Figures~\ref{fig:sample_varis_rescuenet} and~\ref{fig:sample_varis_floodnet} present the corresponding counting results ($R^2$) on \rNet{} and \fNet, respectively.
As detection quality improves with more fine-tuning data, overall framework performance increases in most cases, with particularly notable gains observed for \qwenvl.
Notably, relatively small number of samples ($100$ and $500$) already yields $R^2$ scores comparable to those obtained with the full fine-tuning set.
This result indicates that, across different \acp{lvlm} backbones, the framework remains highly effective under limited data conditions, while consistently outperforming G-DINO*.
However, when the fine-tuning set exceeds $1000$ samples, G-DINO* occasionally surpasses our approach.

Interestingly, G-DINO* without any fine-tuning achieves an $R^2$ of approximately $0.3$ despite very poor detection accuracy ($AP^{75}=0.077$).
Closer inspection reveals that while many predicted bounding boxes are poorly localized, their total count per image roughly matches the true building count.
This suggests that evaluating performance only from metrics focusing on the number of bounding boxes produced by a \ac{cv} model can be unreliable, as these boxes may not correspond to correctly localized objects.
\begin{table}[tb]
    \centering
    \resizebox{0.9\columnwidth}{!}{%
    \begin{tabular}{l|c|c|c|c|c|c||c|c|c|c|c}\toprule
 & \multicolumn{6}{c||}{\textbf{\rNet}}&\multicolumn{5}{c}{\textbf{\fNet}}\\\midrule
          \textbf{\#Fine-tuning samples} &0&  100&  500&  1000& 2000 &4044 & 0& 100& 500& 1000&1895
\\ \hline
          \textbf{G-DINO} &0.499&  0.71&  0.76&  0.793&  0.813&0.839
 & 0.445& 0.772& 0.822& 0.827&0.840
\\
          \textbf{G-DINO*} &0.118&  0.42&  0.54&  0.583&  0.633&0.677
 & 
0.077& 0.655& 0.714& 0.731&0.762
\\ \bottomrule 
    \end{tabular}
    }
    \caption{$AP^{75}$ of G-DINO and G-DINO* on the \rNet~and \fNet~test sets, across varying fine-tuning set sizes.}
    \label{tab:mAP}
\end{table}

\subsubsection{Impact of Mask Representation}
\begin{figure}
    \begin{subfigure}{.39\linewidth}
    \centering
    \includegraphics[width=0.8\linewidth]{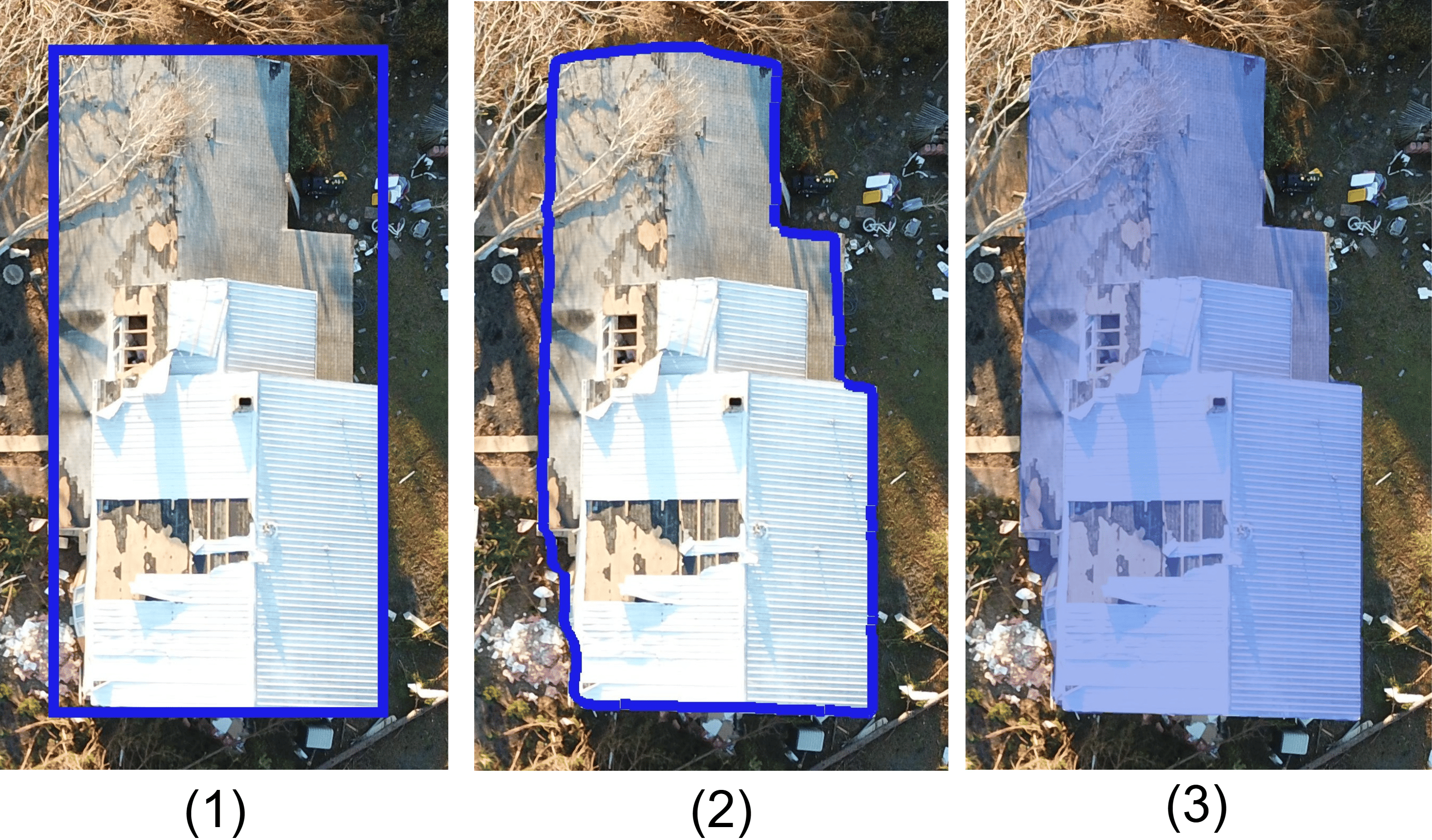}
    \caption{Illustration of the different mask types used for building detection: (1) Rectangular bounding box, (2) polygonal box, and (3) pixel-level segmentation.}
    \label{fig:mask}
    \end{subfigure}
    \hspace{\fill}
    \begin{subfigure}{.49\linewidth}
    \centering
    \includegraphics[width=1.0\linewidth]{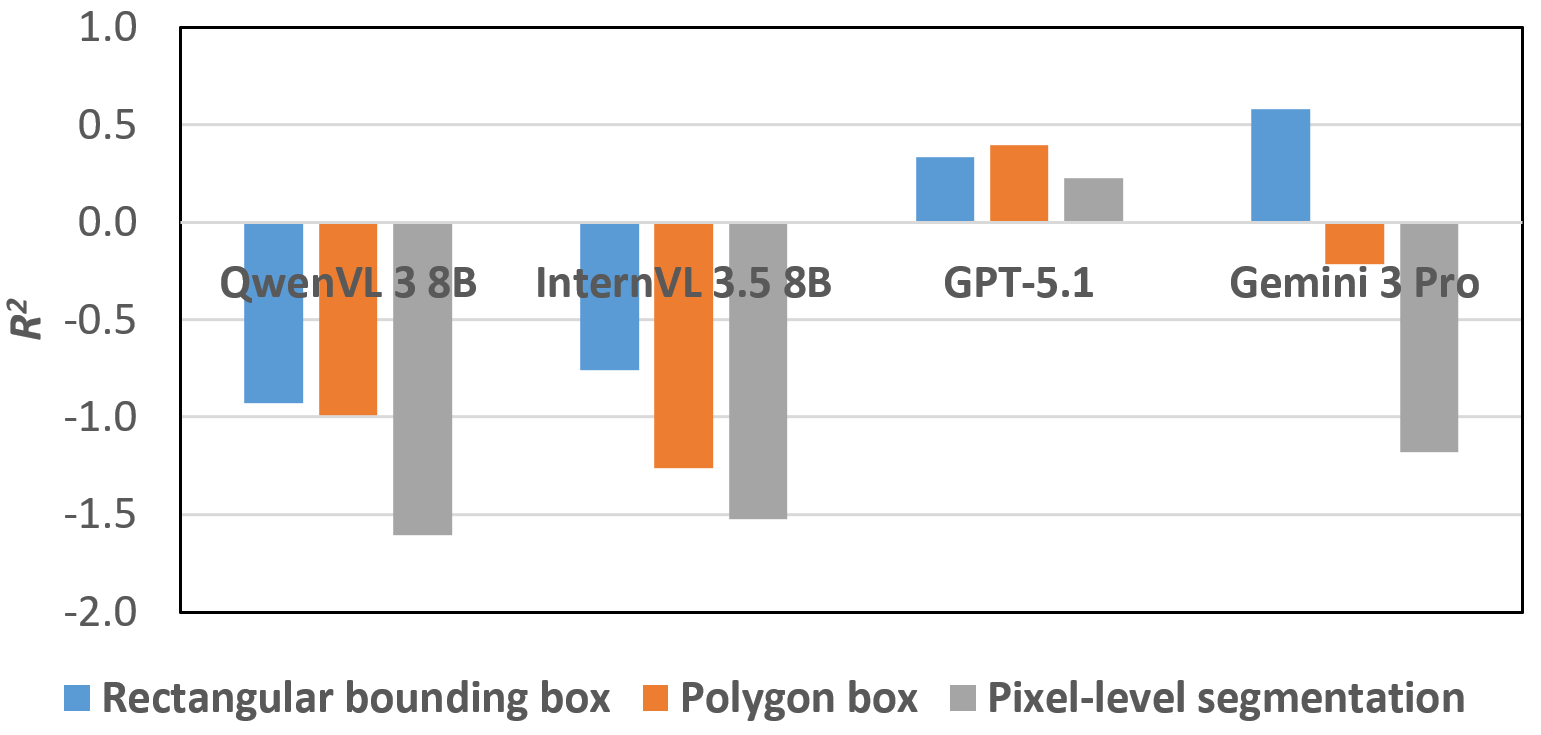}
    \caption{$R^2$ performance of \acp{lvlm} on \rNet~when building identification is done with each mask type.}
    \label{fig:mask_impact}
    \end{subfigure}
    \caption{Impact of mask representation on \ac{lvlm} damage assessment performance. (a) Example of the different detection boxes considered. (b) $R^2$ scores on \rNet~for each \ac{lvlm} when provided with each mask type.}
    \label{fig:exp_mask}
\end{figure}

To investigate how the representation of areas corresponding to a detected building affects \ac{lvlm} reasoning, we evaluate three types of localization mask: rectangular bounding boxes (used throughout this paper), polygonal boxes, and pixel-level segmentation masks.
Figure~\ref{fig:mask} illustrates representative examples of each type. 
To isolate the effect of mask representation, all masks in this experiment are derived from ground-truth annotations.

As shown in Figure~\ref{fig:mask_impact}, mask representation has a significant impact on framework performance.
Rectangular bounding boxes achieve the most stable and consistently strong results across all \acp{lvlm}.
Polygonal masks yield comparable performance in some cases but exhibit notable degradation in others.
Pixel-level segmentation performs worst overall, and in some configurations producing negative $R^2$ values for \gemini.

We attribute this ordering to the visual information preserved by each mask type. Rectangular and polygonal masks retain the full appearance of building regions, allowing \acp{lvlm} to observe structural damage cues directly.
Pixel-level segmentation, however, requires overlaying a colored mask to delineate building boundaries, which can obscure damaged surface details and hinder damage-level interpretation.
These findings suggest that moderate spatial guidance (such as bounding boxes) is more effective than precise segmentation for supporting high-level reasoning in \acp{lvlm}-based damage assessment.

\section{Discussion and Perspectives}
\subsection{Analysis of Failure Cases}

We further investigate the performance degradation discovered when our framework integrates open-source \acp{lvlm} on FloodNet.
A similar trend is observed in the RescueNet dataset for intact building counting (see Table~\ref{tab:intact_rescuenet}). 
We hypothesize that \ac{lvlm}-alone models possess sufficient internal priors to identify objects lacking disaster-related structural distortions, such as intact or flooded buildings without visible damage. 
In these scenarios, the introduction of bounding boxes may become redundant. 
By isolating the object, the bounding box may also strip essential environmental context (e.g., surrounding floodwater or blocked access routes) that the model requires for relational reasoning. 
This suggests that the utility of external detection boxes is non-uniform and highly dependent on the visual processing strategies inherent to the specific \ac{lvlm} architecture.

\subsection{Dataset Characteristics and Limitations}

Both datasets are derived from \ac{uav} imagery captured along continuous flight paths, resulting in significant visual overlap between sequential frames. 
Our investigation reveals that images from the same flight sequence occasionally appear across training, validation, and test splits. 
This introduces the risk of spatial autocorrelation and potential data contamination, as the model may encounter visually familiar regions at test time.
While we acknowledge that this is a limitation common to many aerial datasets, rigorously quantifying the impact of this sequence leakage on model performance remains an important direction for future work.

Both datasets are geographically constrained to specific regions of the United States, which may limit the cross-regional transferability of our findings. 
Models optimized for particular architectural styles or environmental conditions often struggle when deployed in regions with different building morphologies or disaster management infrastructures.
Our framework is not exempt from this concern.

We identify a significant discrepancy between the formal definition of a flooded building in \fNet{} and its ground-truth annotations.
Although the dataset defines a flooded building by direct water contact (at least one side touching floodwater), many annotations appear to follow an operational logic, labeling buildings as flooded if their access routes are submerged. 
While pragmatically useful for emergency response, this inconsistency introduce semantic noise for prompt-based models like \gdino and \acp{lvlm}, which rely on strict textual-visual alignment. 
This mismatch between linguistic definitions and visual labels likely undermines both zero-shot inference and the broader generalizability of the framework.

\begin{table*}[t]
\centering
\small
\resizebox{0.9\columnwidth}{!}{%
\begin{tabular}{l|c|rr|rr|rr|rr}
\toprule
\multirow{2}{*}{\textbf{Method}} & 
\multirow{2}{*}{\textbf{Fine-tuning}}& 
\multicolumn{2}{c}{\textbf{\qwenvl}}& 
\multicolumn{2}{|c}{\textbf{\internvl}}& 
\multicolumn{2}{|c}{\textbf{\gpt}}& 
\multicolumn{2}{|c}{\textbf{\gemini}}\\
&  & RMSE ($\downarrow$) & R$^2$ ($\uparrow$) 
  & RMSE ($\downarrow$) & R$^2$ ($\uparrow$) 
  & RMSE ($\downarrow$) & R$^2$ ($\uparrow$) 
  & RMSE ($\downarrow$) & R$^2$ ($\uparrow$)  \\
\midrule
\textbf{\ac{lvlm}}& \textcolor{lightgray}{\xmark}& 1.109& 0.494 & 1.317& 0.287 & 1.255& 0.352 & 1.555& 0.007
 \\\midrule

\multirow{3}{*}{\textbf{\ac{lvlm}+G-DINO}} & {Not FT} & 1.449& 0.137 & 1.432& 0.152 & 1.225& 0.383 & 1.148& 0.458
 \\

{} & {FT-100}& 1.514& 0.058 & 1.428& 0.162 & 1.235& 0.373 & 1.081& 0.520
 \\

{} & {FT-Full} & 1.467& 0.116 & 1.386& 0.208 & 1.245& 0.363 & 1.105& 0.498
 \\\midrule

\textbf{\ac{lvlm}+\textit{GTBBox}} & \textcolor{lightgray}{\xmark}& 1.259& 0.348 & 1.337& 0.262 & 1.272& 0.334 & 1.034& 0.560
 \\

\bottomrule
\end{tabular}
}
\caption{Performance of our framework on \rNet~for counting intact buildings, illustrating cases where combining G-DINO with \acp{lvlm} does not improve over the \ac{lvlm}-alone baseline. ``\textit{GTBBox}'' represents perfect bounding boxes extracted from ground-truth. ``Not FT'', ``FT-100'' and ``FT-Full'' denote G-DINO without fine-tuning, fine-tuned with 100 samples, and fine-tuned with the full dataset, respectively.}
\label{tab:intact_rescuenet}
\end{table*}

\subsection{Future Directions}
Finally, several constraints of the current study merit further exploration. 
First, while we focused on training-free integration to assess the baseline reasoning of \acp{lvlm}, future benchmarks should include comparisons with models fine-tuned on the target domain. 
Second, our framework is agnostic to the choice of \ac{cv} detector; however, we have yet to evaluate it against fully supervised models like YOLO~\cite{sapkota2025yolo26}. 
Given that supervised detectors require substantial high-quality annotated data to outperform open-vocabulary alternatives, future research will focus on training dedicated detectors on expanded datasets. This will provide clearer architectural guidelines for practitioners implementing automated systems for post-disaster building damage assessment.

\section{Conclusion}

In this paper, we propose a hybrid framework for post-disaster building damage assessment that decouples building detection from damage assessment by combining the localization capabilities of \ac{cv} models with the reasoning abilities of \acp{lvlm}. 
A \ac{cv}-based detection module identifies buildings and generates bounding boxes, while the reasoning module based on \ac{lvlm} classifies damage levels and provides contextual interpretation. 
In this work, we use \gdino as the \ac{cv} model.

Evaluated on RescueNet and FloodNet, the framework consistently outperforms isolated baselines, particularly under low-data conditions. 
Ablation studies further show that increasing the number of fine-tuning samples for \gdino improves the overall performance, and that rectangular bounding boxes perform better than polygonal boxes and pixel-level segmentation masks. 
Finally, we discuss failure cases, dataset limitations, and directions for future work.

\bibliographystyle{splncs04}
\bibliography{reference}

\appendix
\clearpage
\section{Appendix}
\subsection{Further Analysis on Datasets}
\begin{figure}
    \centering
    \includegraphics[width=\linewidth]{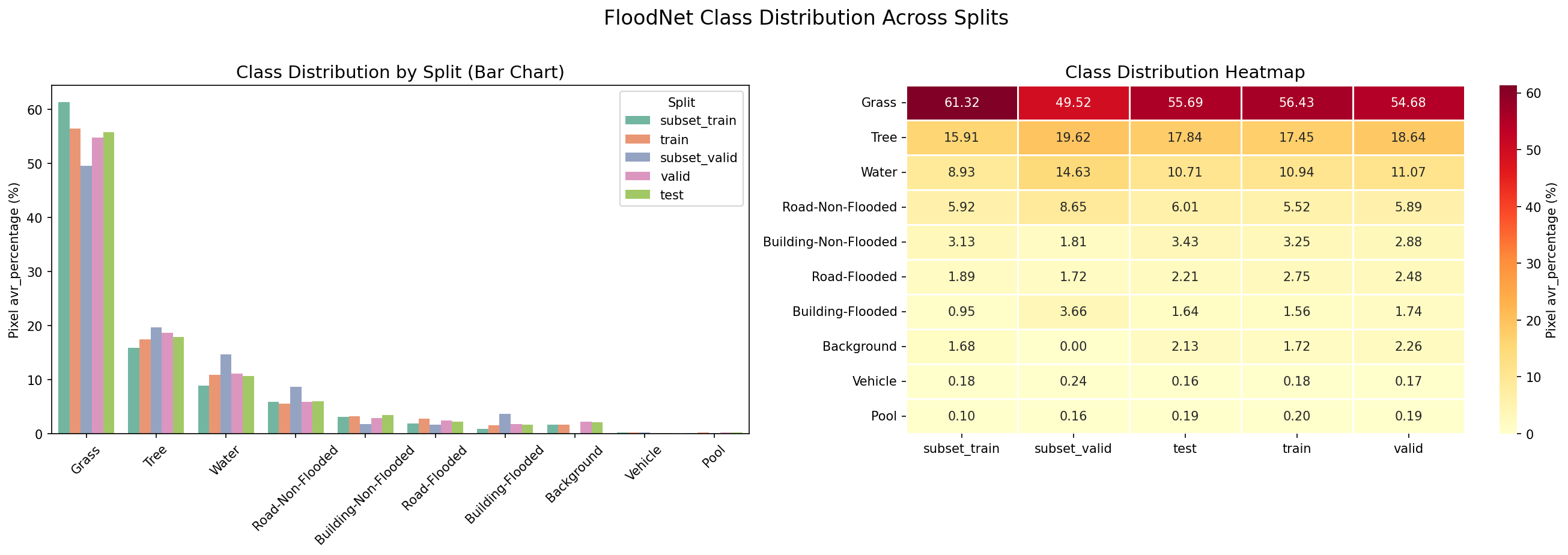}
    \caption{Distribution of average pixel occupancy per image across \fNet{} data splits and the FT-100 fine-tuning subset (subset\_train, subset\_valid), shown as class-wise percentages. Classes are imbalanced in terms of average pixel occupancy per image and the FT-100 sampling strategy successfully preserves the original class distribution.}
    \label{fig:floodnet_pixelaverage_distri}
\end{figure}

\begin{figure}
    \centering
    \includegraphics[width=\linewidth]{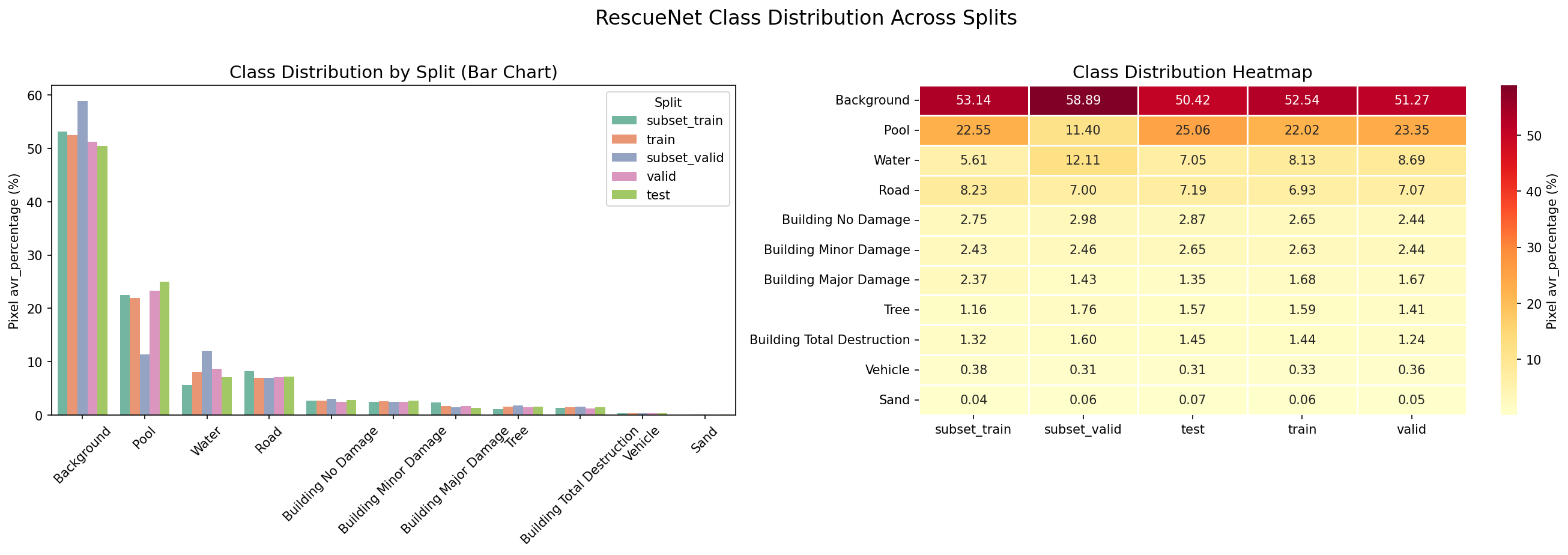}
    \caption{Distribution of average pixel occupancy per image across \rNet{} data splits and the FT-100 fine-tuning subset (subset\_train, subset\_valid), shown as class-wise percentages. Classes are imbalanced in terms of average pixel occupancy per image and the FT-100 sampling strategy successfully preserves the original class distribution.}
    \label{fig:rescuenet_pixelaverage_distri}
\end{figure}

Figures~\ref{fig:floodnet_pixelaverage_distri} and~\ref{fig:rescuenet_pixelaverage_distri} represent the distribution of the average pixel occupancy of each class across data splits for \fNet{} and \rNet, respectively.
Both datasets exhibit strong class imbalance at the pixel level: non-building classes dominate the pixel count: \emph{Grass}, \emph{Tree}, \emph{Water} account for over $70$\% of pixels in \fNet{}, while \emph{Background}, \emph{Pool}, \emph{Water} together account for over $80$\% in \rNet.
In contrast, building-related classes occupy only a small fraction of total pixels: Building-Flooded and Building-Non-Flooded together account for under $6$\% in \fNet, and all four \rNet~damage classes combined (No Damage, Minor, Major, Total Destruction) account for under $9$\%.

This imbalance is particularly problematic for building damage assessment. 
Accurately classifying damage level depends on fine-grained visual cues such as roof condition, structural collapse, or surrounding debris.
Therefore, a low pixel share means models are trained and evaluated on comparatively little direct evidence for the classes that matter most. 
Background-dominated images can bias models toward features unrelated to buildings, while the limited pixel footprint of building regions, leaves little visual detail to distinguish between damage levels. 
Moreover, standard pixel-level metrics can be misleading under such imbalance, as high apparent accuracy may largely reflect correct classification of dominant background regions rather than reliable building-level discrimination. 
This motivates the building-centric counting evaluation adopted in our framework, which directly targets these underrepresented but critical classes rather than relying on aggregate pixel-level metrics that can mask poor building-specific performance.

\begin{figure}
    \centering
    \includegraphics[width=0.7\linewidth]{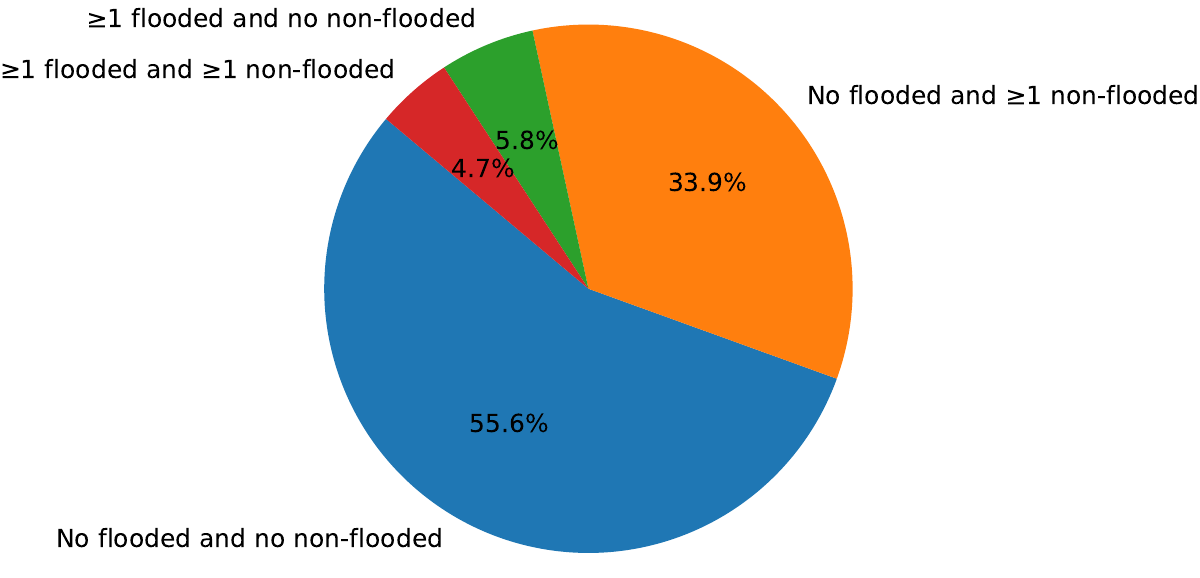}
    \caption{Distribution of samples with existences of flooded and non-flooded buildings on the FloodNet test set.}
    \label{fig:ratio_floodnet}
\end{figure}

Figure~\ref{fig:ratio_floodnet} presents the distribution of building composition across \fNet~test images, based on the presence of flooded and non-flooded buildings. 
Approximately $90$\% of images contain zero flooded buildings at all ($55.6$\% with no buildings of either type, and $33.9$\% with only non-flooded buildings), confirming a strong class imbalance toward the non-flooded case.
The most challenging case for damage assessment is when images contain a mix of flooded and non-flooded buildings.
For these images, a model must correctly discriminate between damage levels within the same scene.
However, in \fNet~these images represent only $4.7$\% of the dataset. 

This imbalance raises an important concern for evaluation: a model that simply detects the presence of floodwater in an image and labels all buildings accordingly (without performing per-building discrimination) could achieve misleadingly high accuracy. 
This underscores the importance of evaluation protocols, such as ours, that assess performance at the level of individual buildings rather than relying on image-level or scene-level, which can substantially overstate real-world model capability on imbalanced datasets like \fNet.

\newpage

\subsection{Scatter plot of the counting task}

\begin{figure*}[htpb]
	\centering
    \begin{subfigure}{0.30\linewidth}
		\includegraphics[width=\linewidth]{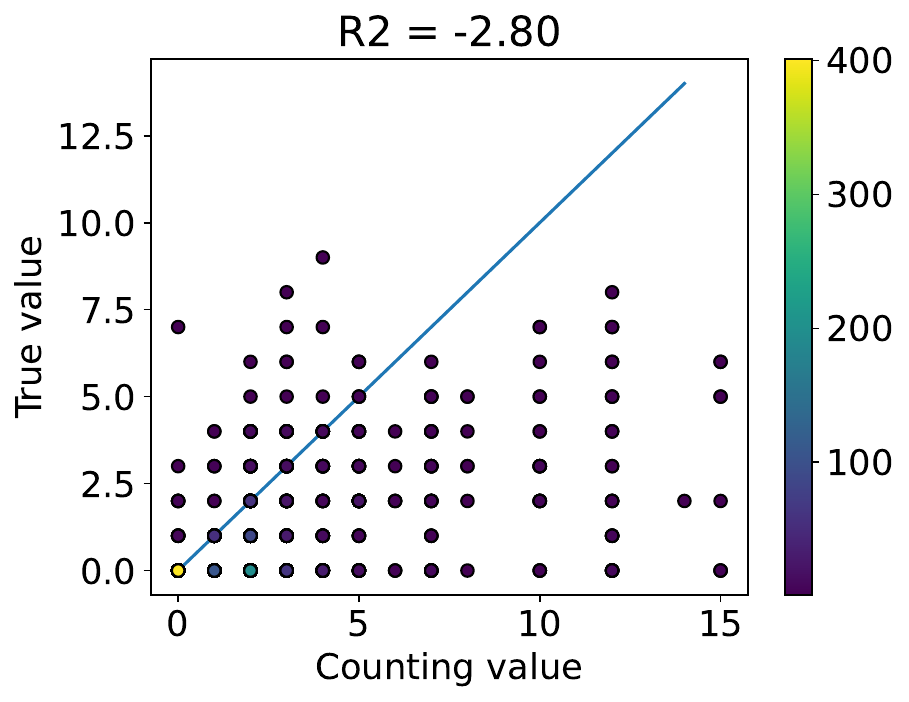}
		\caption{\qwenvl only}
    \end{subfigure} 
    \begin{subfigure}{0.30\linewidth}
		\includegraphics[width=\linewidth]{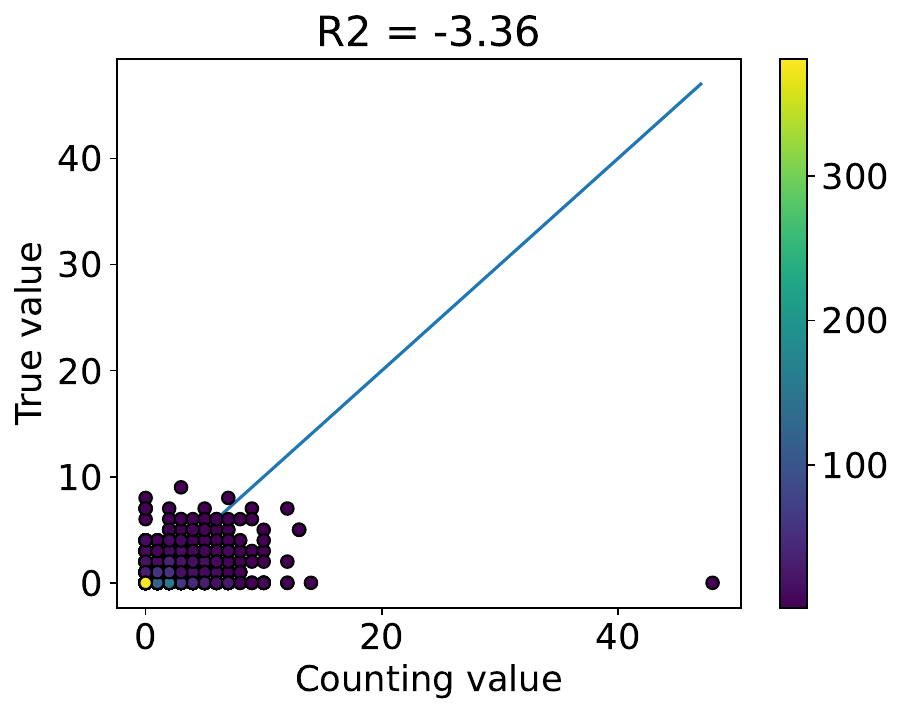}
		\caption{\qwenvl+ G-DINO (Not FT)}
    \end{subfigure}\\
    \begin{subfigure}{0.30\linewidth}
    		\includegraphics[width=\linewidth]{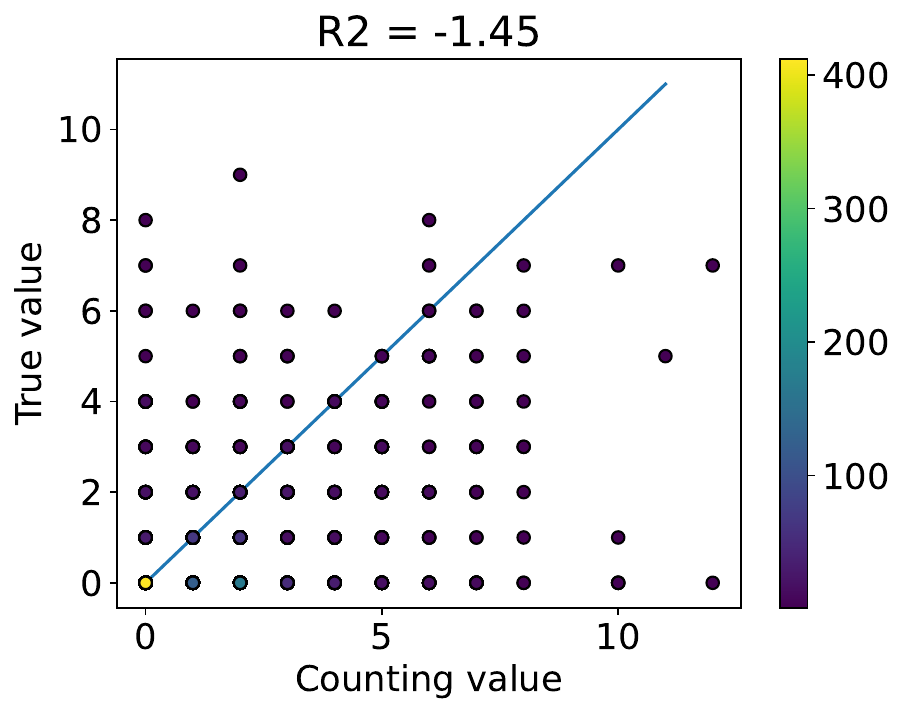}
    		\caption{\qwenvl+ G-DINO (FT-100)}
    \end{subfigure}
    \begin{subfigure}{0.30\linewidth}
    		\includegraphics[width=\linewidth]{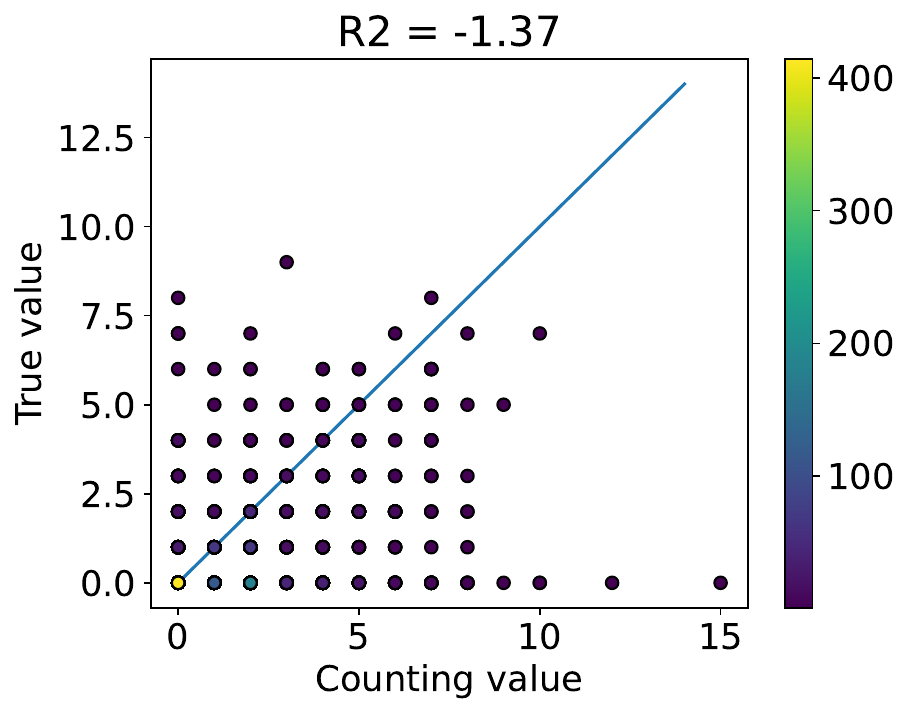}
    		\caption{\qwenvl+ G-DINO (FT-FULL)}
    \end{subfigure}
    \begin{subfigure}{0.30\linewidth}
    		\includegraphics[width=\linewidth]{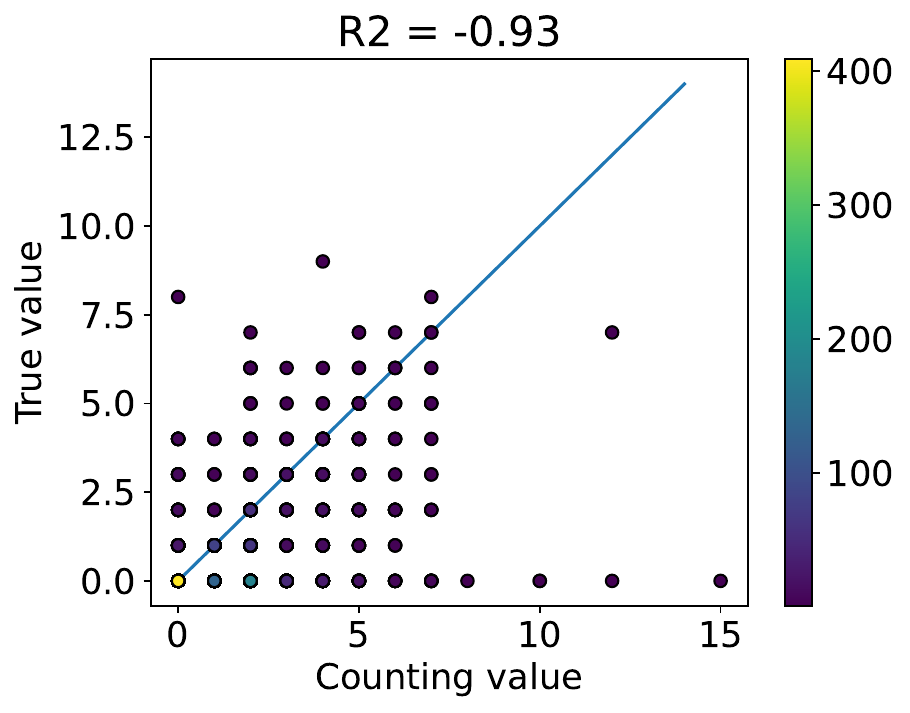}
    		\caption{\qwenvl+True bbox}
    \end{subfigure}
    \caption{Scatter plots of actual counts and counts from models on the RescueNet testing set. Blue lines are used to highlight points with correct counts. To address the overlap of numerous points, color gradients are applied to represent density, providing a clearer visual distinction.}
    \label{fig:scatter_plot_qwenvl}
\end{figure*}

\begin{figure*}[htpb]
	\centering
    \begin{subfigure}{0.30\linewidth}
		\includegraphics[width=\linewidth]{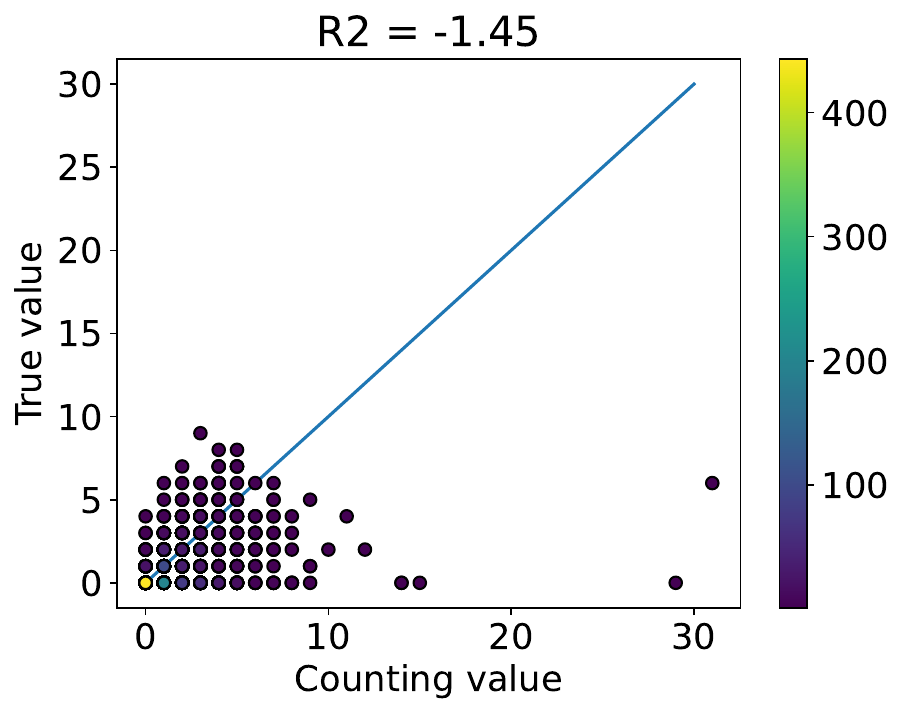}
		\caption{\internvl only}
    \end{subfigure} 
    \begin{subfigure}{0.30\linewidth}
		\includegraphics[width=\linewidth]{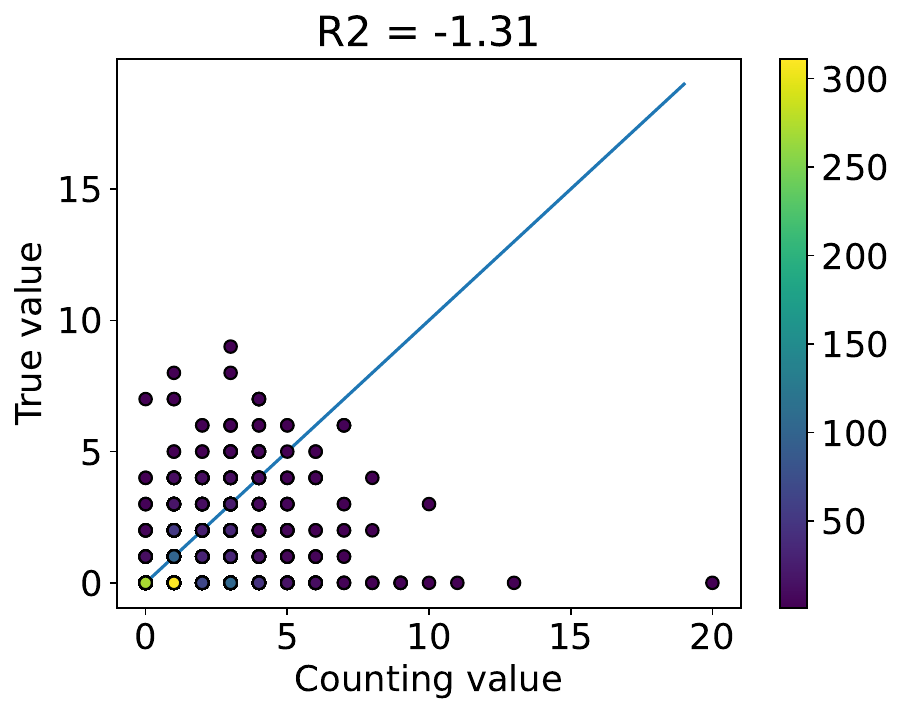}
		\caption{\internvl+ G-DINO (Not FT)}
    \end{subfigure}\\
    \begin{subfigure}{0.30\linewidth}
    		\includegraphics[width=\linewidth]{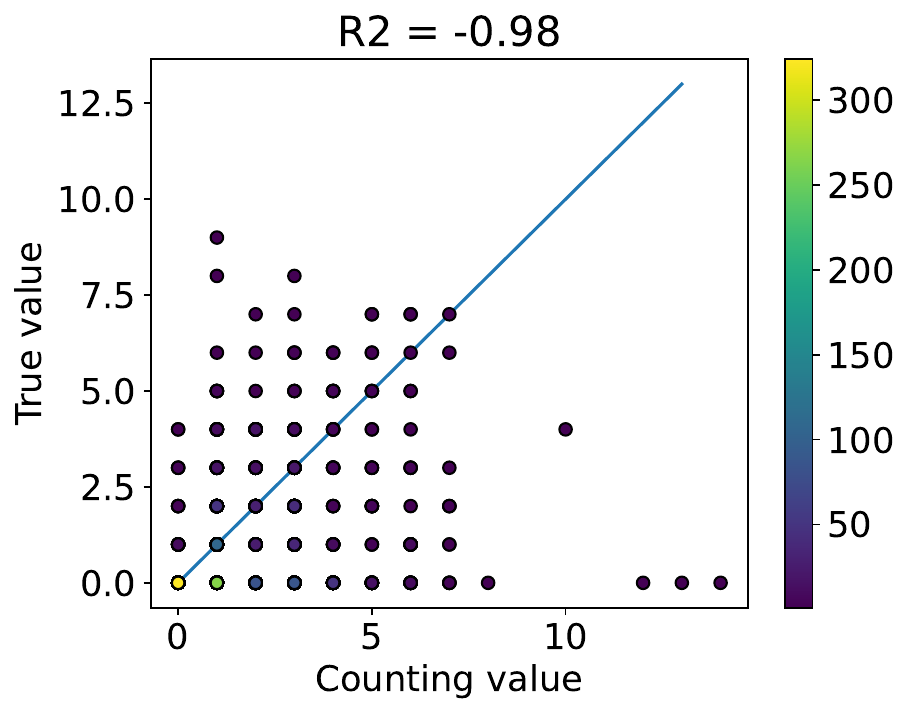}
    		\caption{\internvl+ G-DINO (FT-100)}
    \end{subfigure}
    \begin{subfigure}{0.30\linewidth}
    		\includegraphics[width=\linewidth]{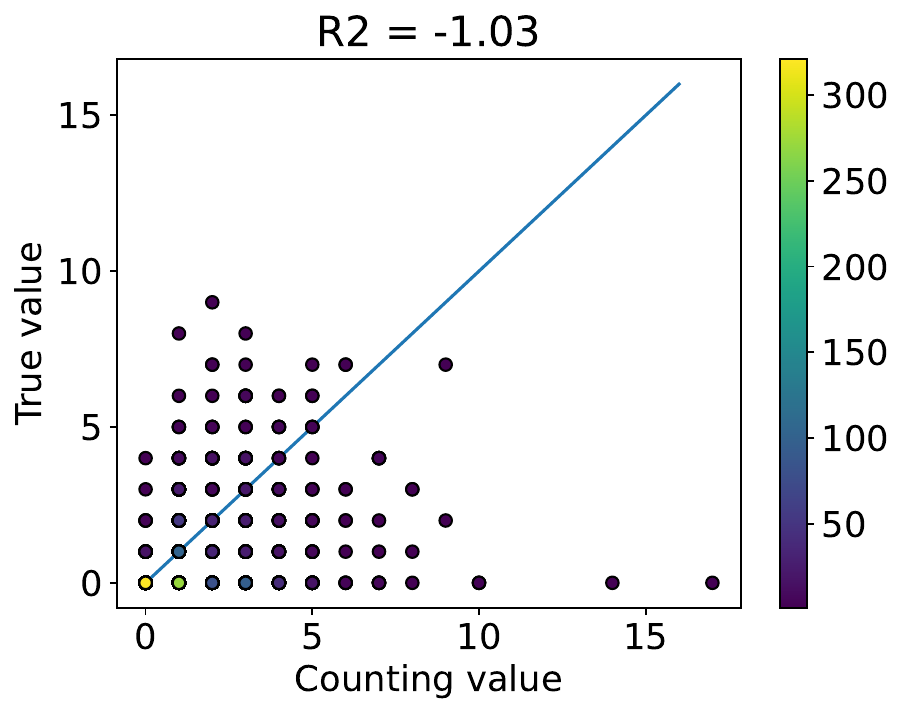}
    		\caption{\internvl+ G-DINO (FT-FULL)}
    \end{subfigure}
    \begin{subfigure}{0.30\linewidth}
    		\includegraphics[width=\linewidth]{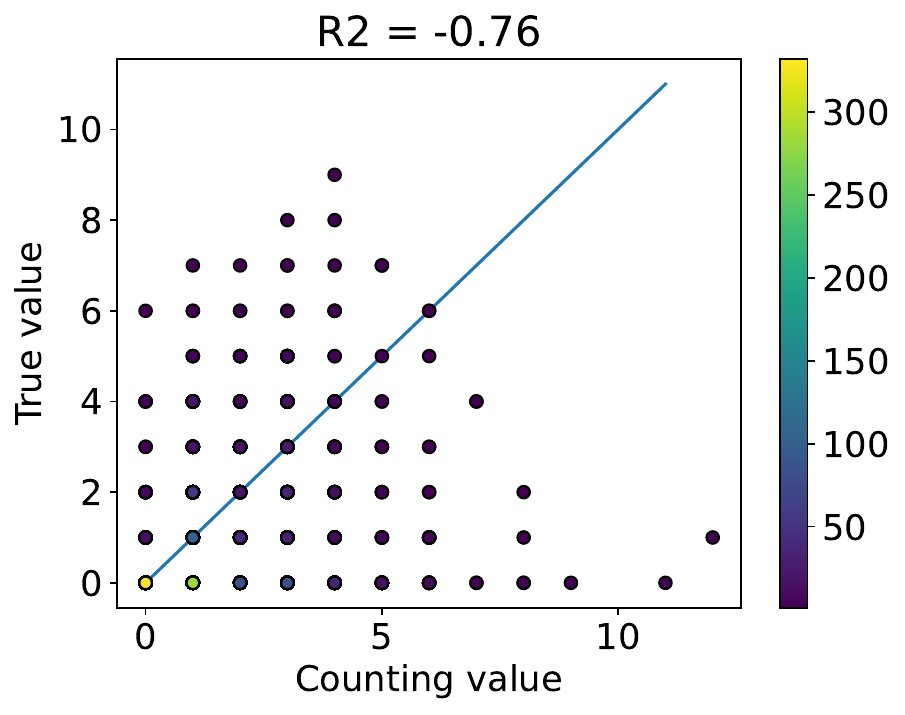}
    		\caption{\internvl+True bbox}
    \end{subfigure}
    \caption{Scatter plots of actual counts and counts from models on the RescueNet testing set.}
    \label{fig:scatter_plot_internvl}
\end{figure*}

\begin{figure*}[htpb]
	\centering
    \begin{subfigure}{0.30\linewidth}
		\includegraphics[width=\linewidth]{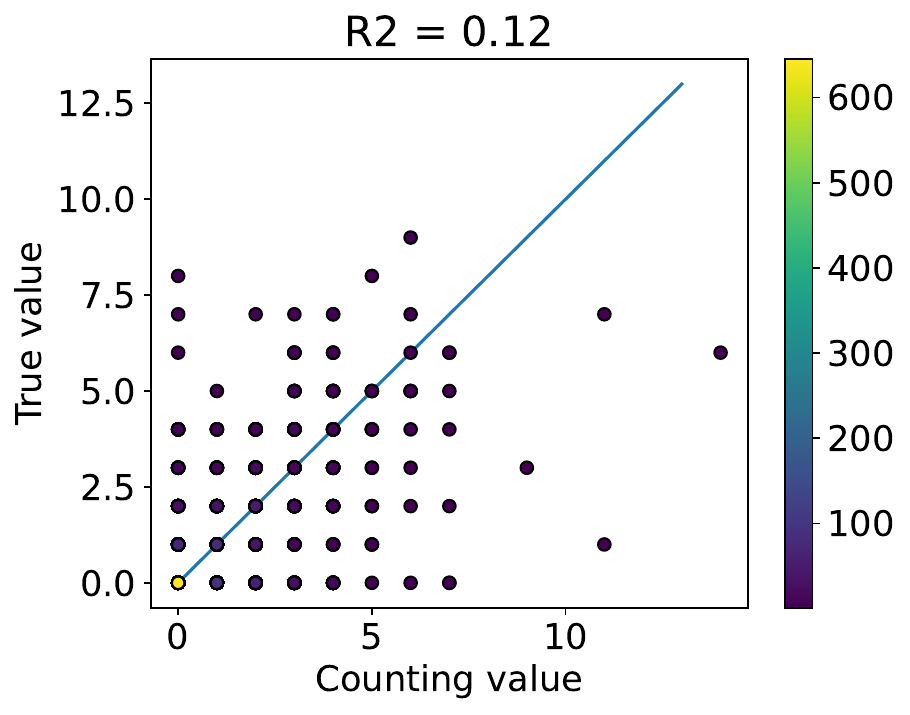}
		\caption{\gpt only}
    \end{subfigure} 
    \begin{subfigure}{0.30\linewidth}
		\includegraphics[width=\linewidth]{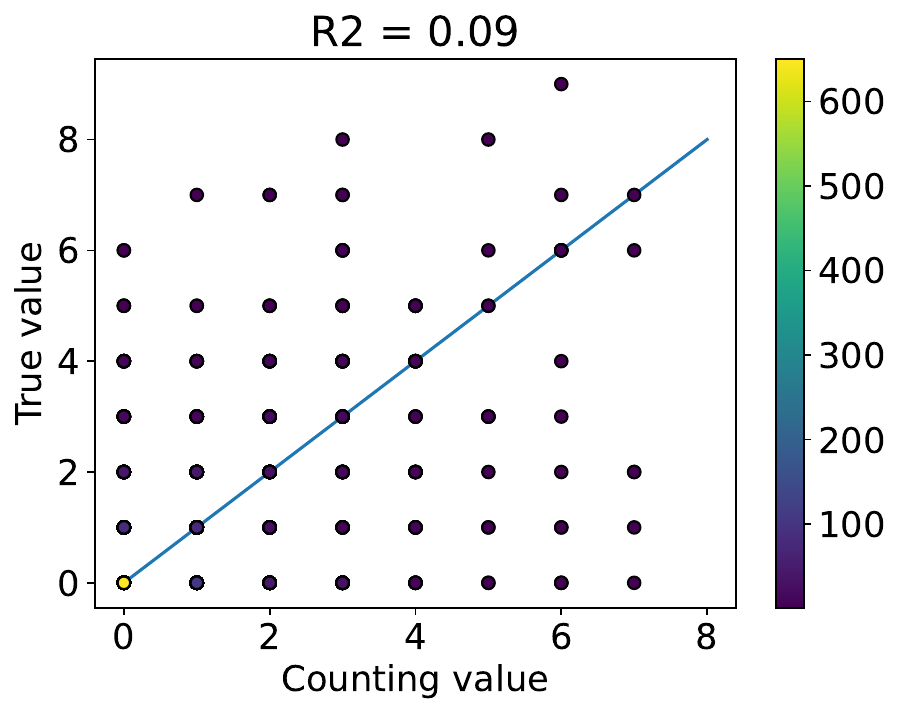}
		\caption{\gpt+ G-DINO (Not FT)}
    \end{subfigure}\\
    \begin{subfigure}{0.30\linewidth}
    		\includegraphics[width=\linewidth]{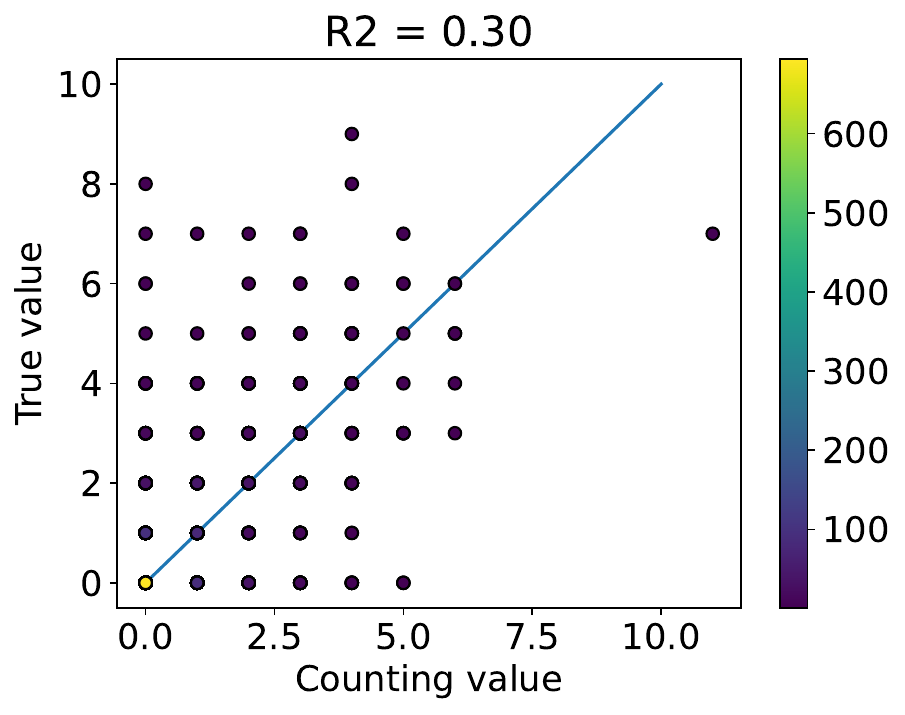}
    		\caption{\gpt+ G-DINO (FT-100)}
    \end{subfigure}
    \begin{subfigure}{0.30\linewidth}
    		\includegraphics[width=\linewidth]{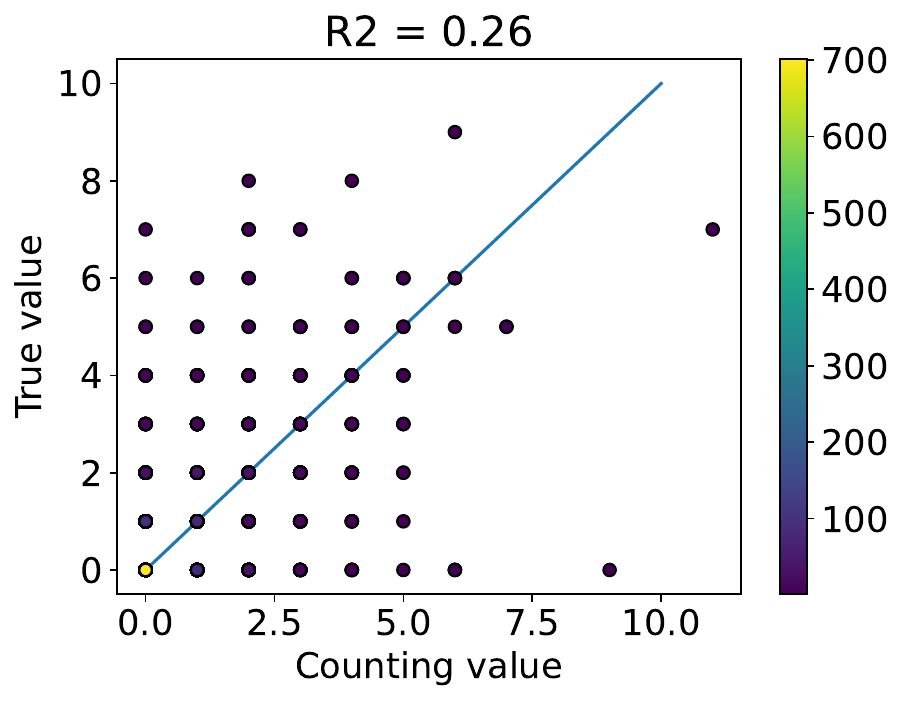}
    		\caption{\gpt+ G-DINO (FT-FULL)}
    \end{subfigure}
    \begin{subfigure}{0.30\linewidth}
    		\includegraphics[width=\linewidth]{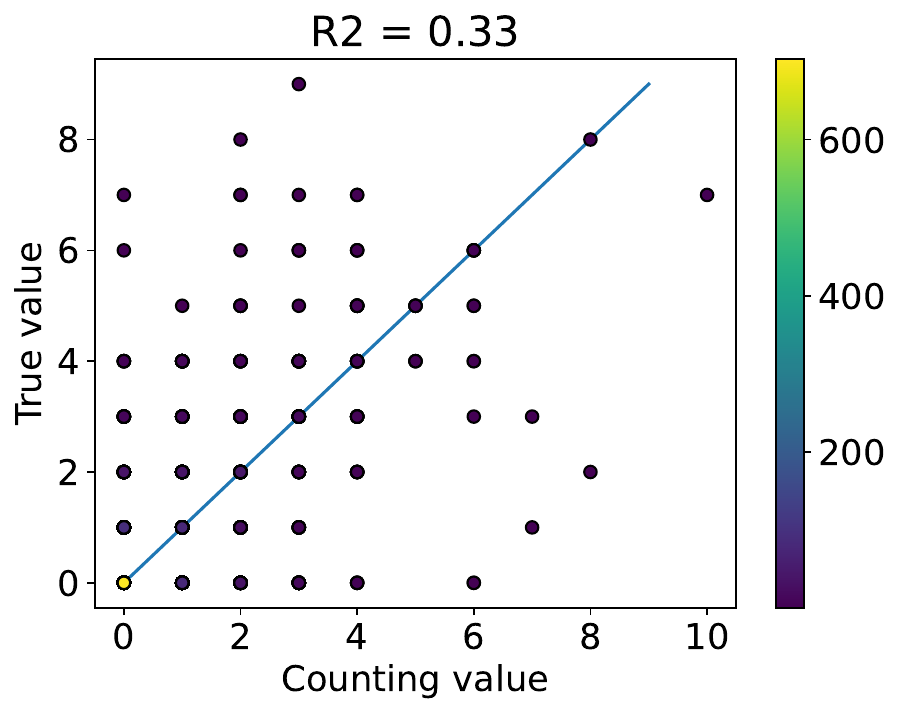}
    		\caption{\gpt+True bbox}
    \end{subfigure}
    \caption{Scatter plots of actual counts and counts from models on the RescueNet testing set. Blue lines are used to highlight points with correct counts. To address the overlap of numerous points, color gradients are applied to represent density, providing a clearer visual distinction.}
    \label{fig:scatter_plot_gpt}
\end{figure*}

\begin{figure*}[htpb]
	\centering
    \begin{subfigure}{0.30\linewidth}
		\includegraphics[width=\linewidth]{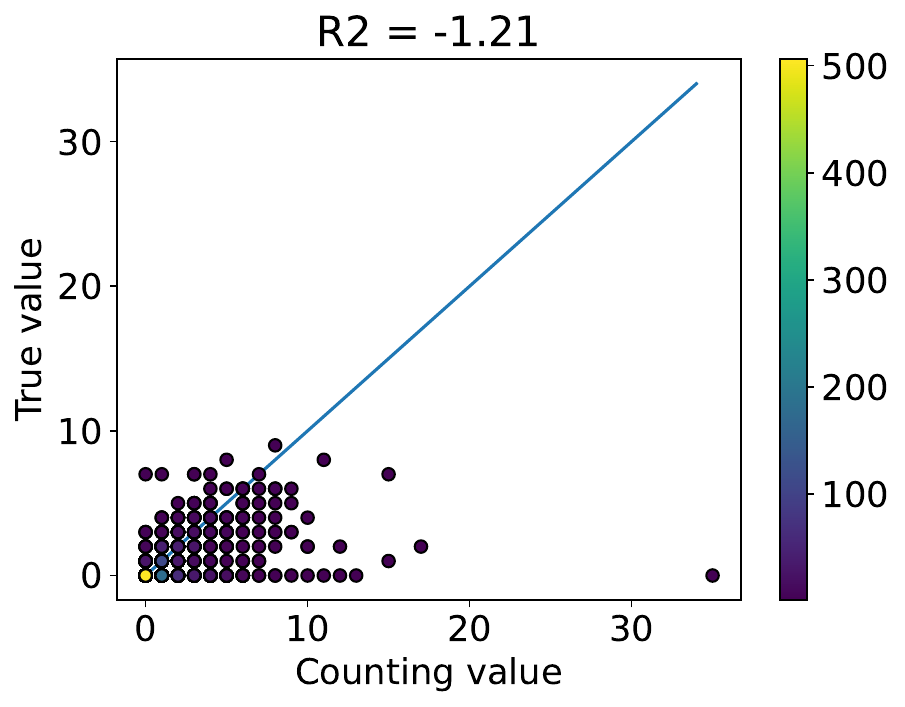}
		\caption{\gemini only}
    \end{subfigure} 
    \begin{subfigure}{0.30\linewidth}
		\includegraphics[width=\linewidth]{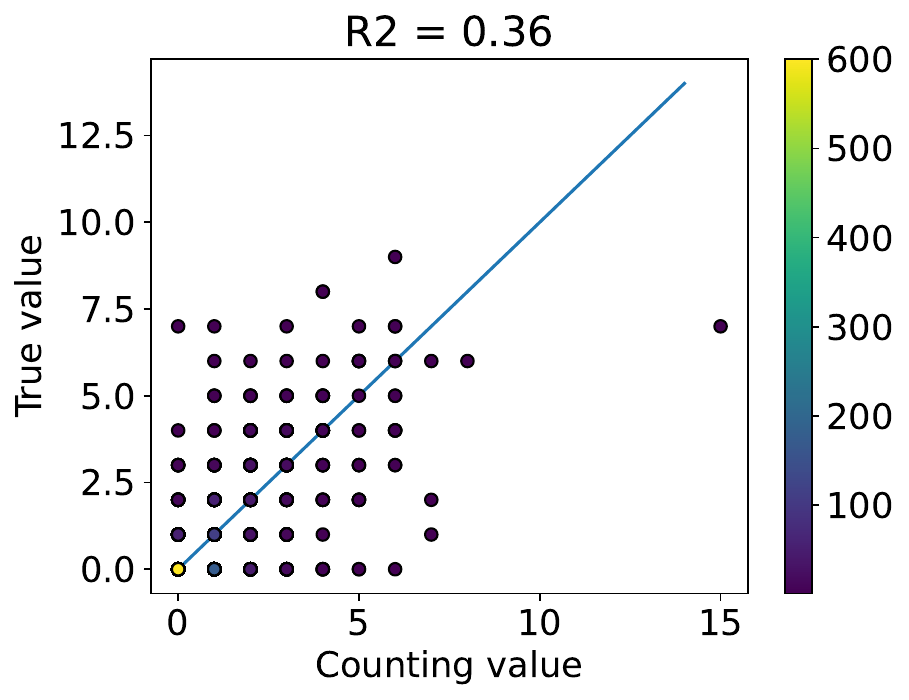}
		\caption{\gemini+ G-DINO (Not FT)}
    \end{subfigure}\\
    \begin{subfigure}{0.30\linewidth}
    		\includegraphics[width=\linewidth]{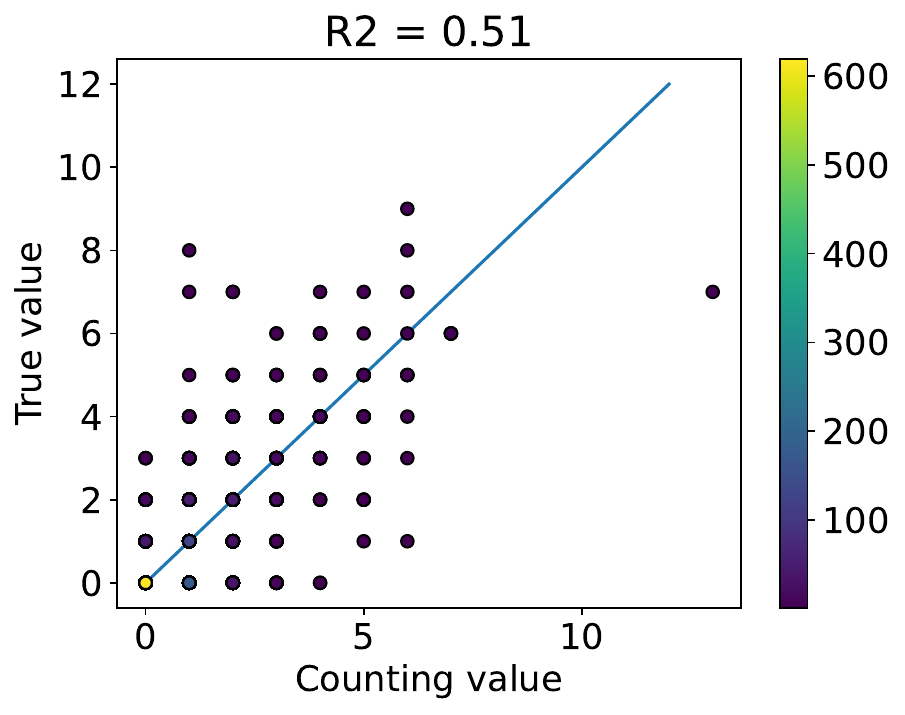}
    		\caption{\gemini+ G-DINO (FT-100)}
    \end{subfigure}
    \begin{subfigure}{0.30\linewidth}
    		\includegraphics[width=\linewidth]{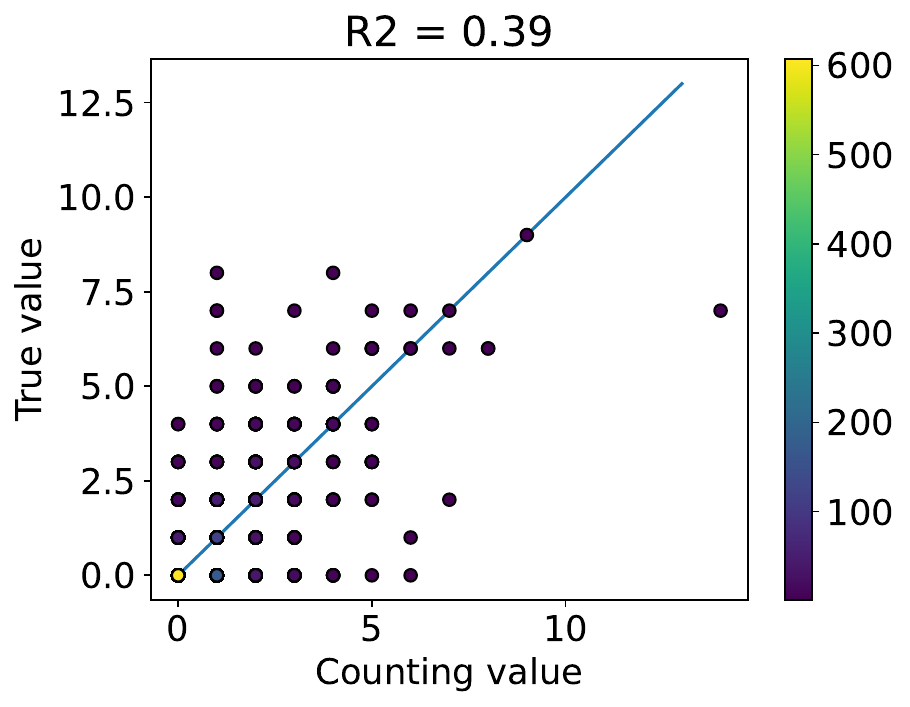}
    		\caption{\gemini+ G-DINO (FT-FULL)}
    \end{subfigure}
    \begin{subfigure}{0.30\linewidth}
    		\includegraphics[width=\linewidth]{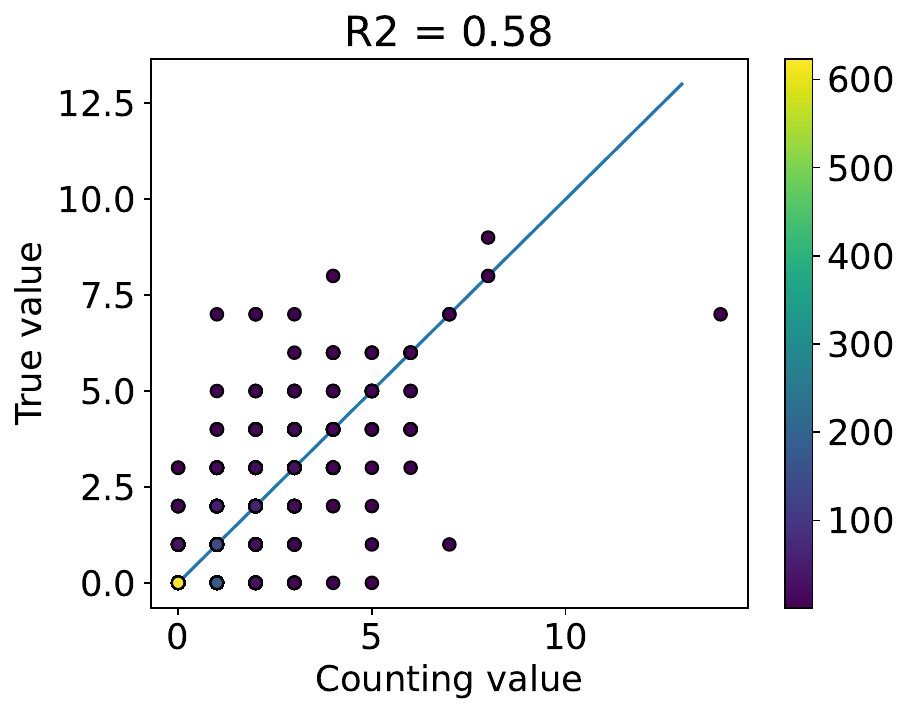}
    		\caption{\gemini+True bbox}
    \end{subfigure}
    \caption{Scatter plots of actual counts and counts from models on the RescueNet testing set.}
    \label{fig:scatter_plot_gemini}
\end{figure*}

\begin{figure*}[htpb]
	\centering
    \begin{subfigure}{0.30\linewidth}
		\includegraphics[width=\linewidth]{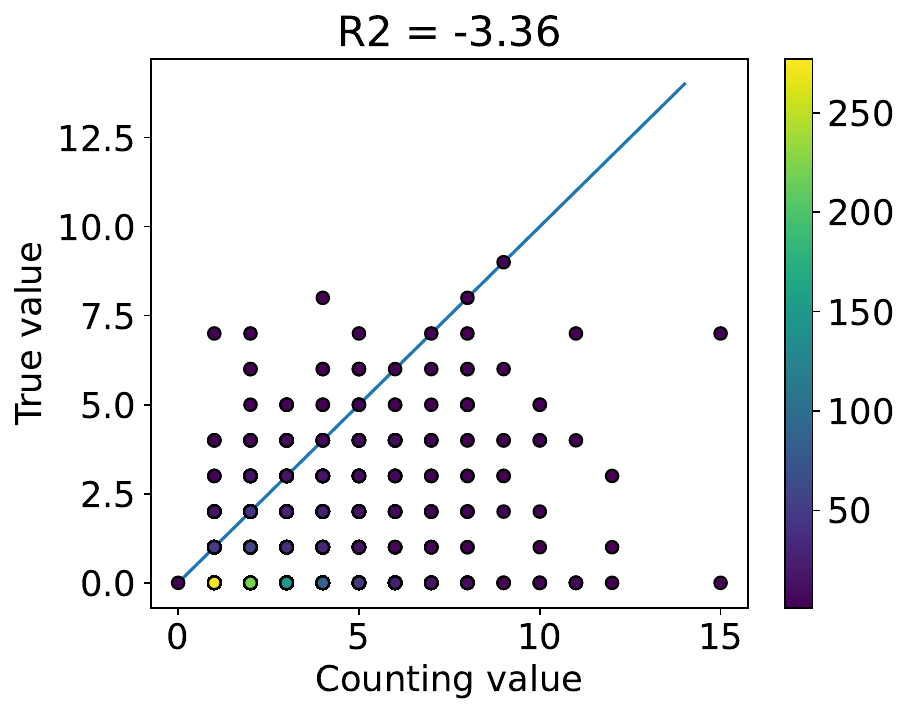}
		\caption{G-DINO* with Not FT}
    \end{subfigure}
    \begin{subfigure}{0.30\linewidth}
    		\includegraphics[width=\linewidth]{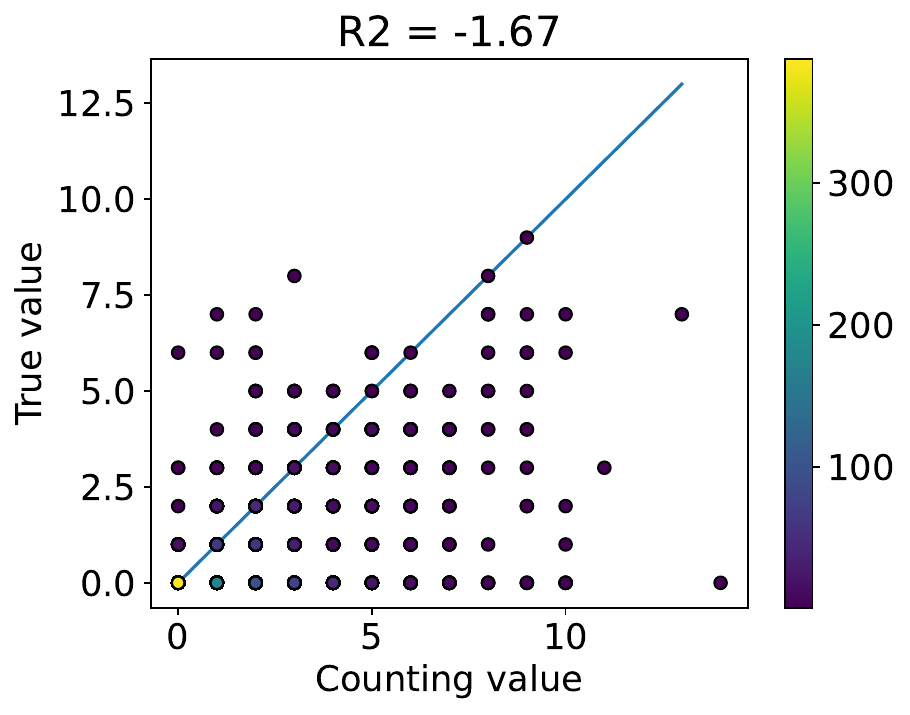}
    		\caption{G-DINO* with FT-100}
    \end{subfigure}
    \begin{subfigure}{0.30\linewidth}
    		\includegraphics[width=\linewidth]{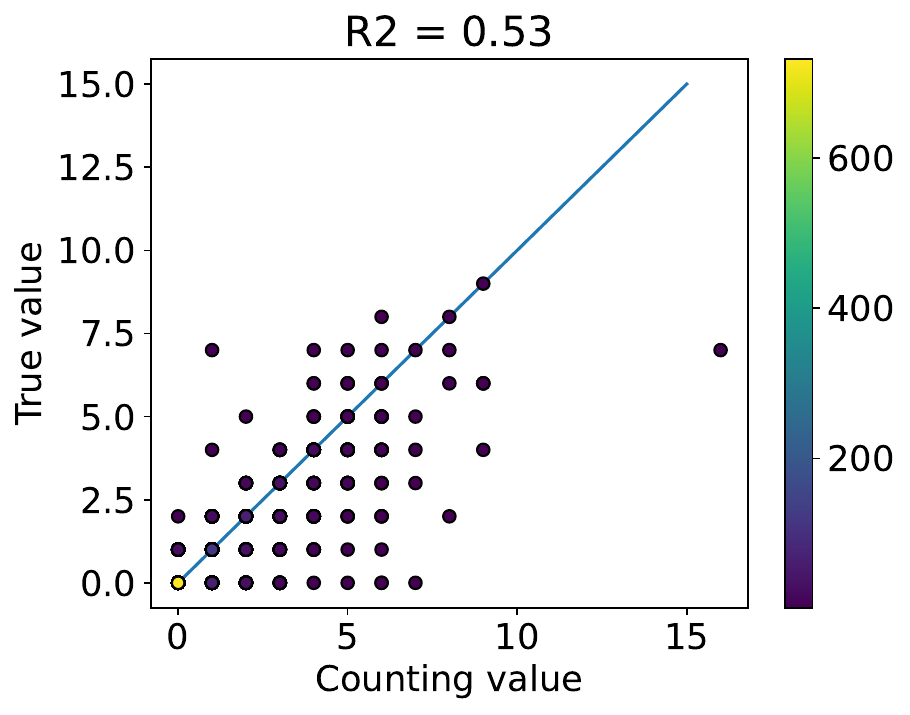}
    		\caption{G-DINO* with FT-FULL}
    \end{subfigure}
    \caption{Scatter plots of actual counts and counts from models on the RescueNet testing set.}
    \label{fig:scatter_plot_gdino}
\end{figure*}

\begin{figure*}[htpb]
	\centering
    \begin{subfigure}{0.30\linewidth}
		\includegraphics[width=\linewidth]{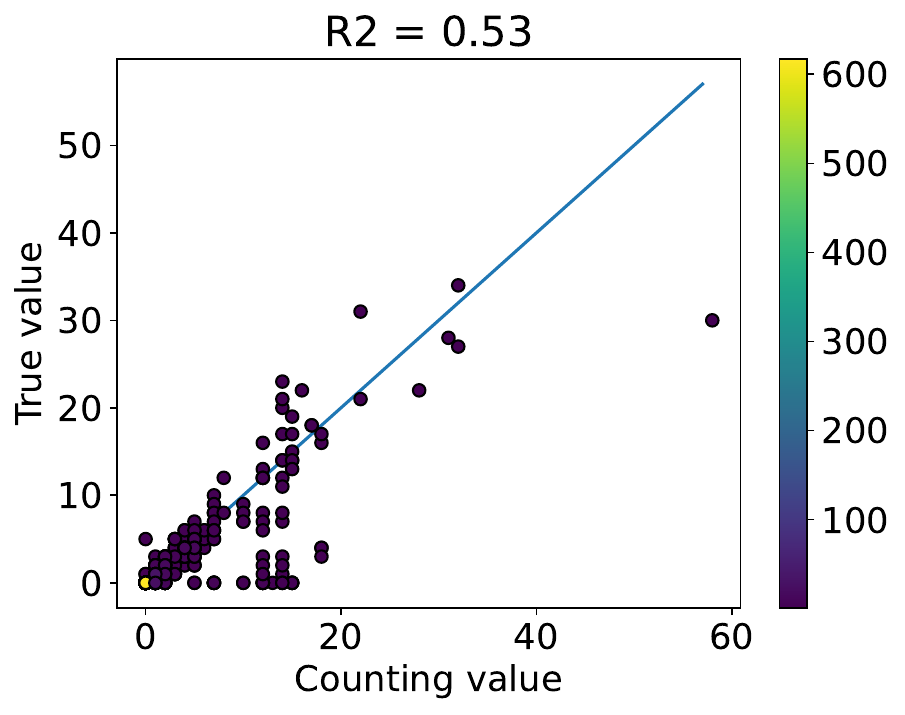}
		\caption{\qwenvl only}
    \end{subfigure} 
    \begin{subfigure}{0.30\linewidth}
		\includegraphics[width=\linewidth]{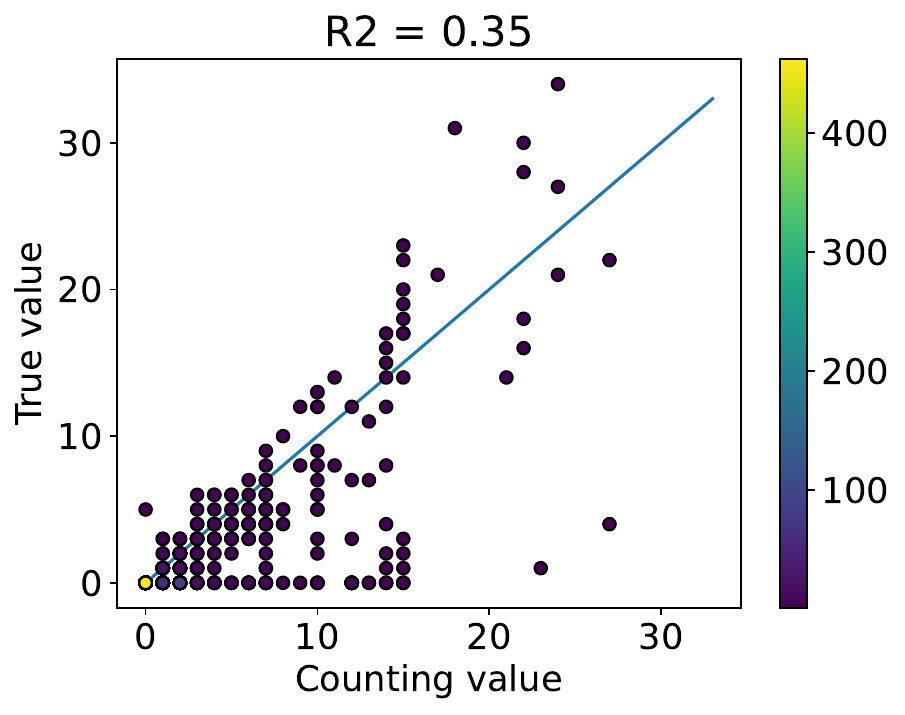}
		\caption{\qwenvl+ G-DINO (Not FT)}
    \end{subfigure}\\
    \begin{subfigure}{0.30\linewidth}
    		\includegraphics[width=\linewidth]{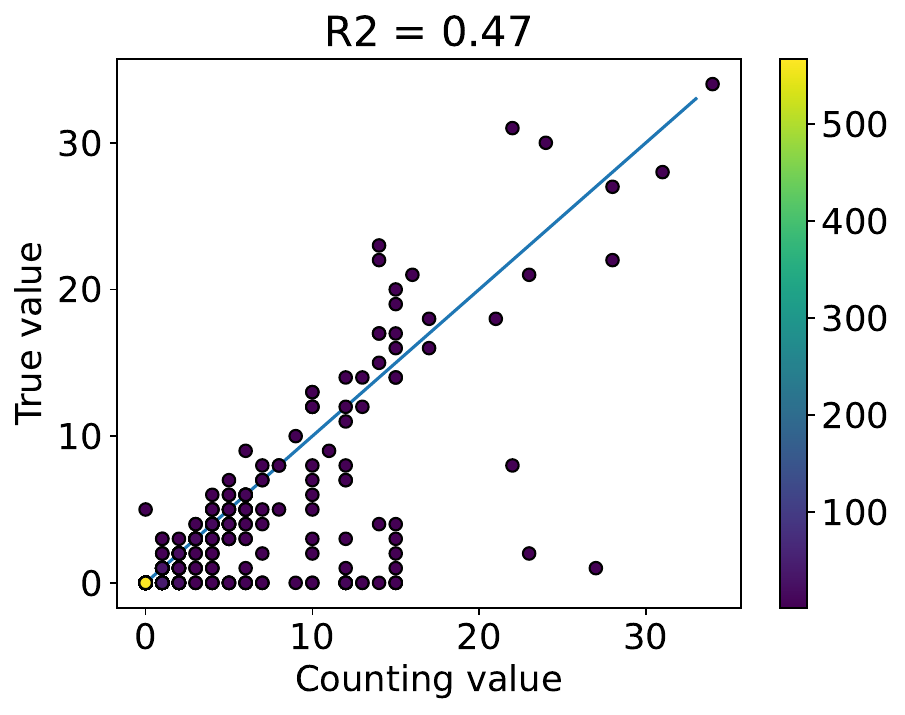}
    		\caption{\qwenvl+ G-DINO (FT-100)}
    \end{subfigure}
    \begin{subfigure}{0.30\linewidth}
    		\includegraphics[width=\linewidth]{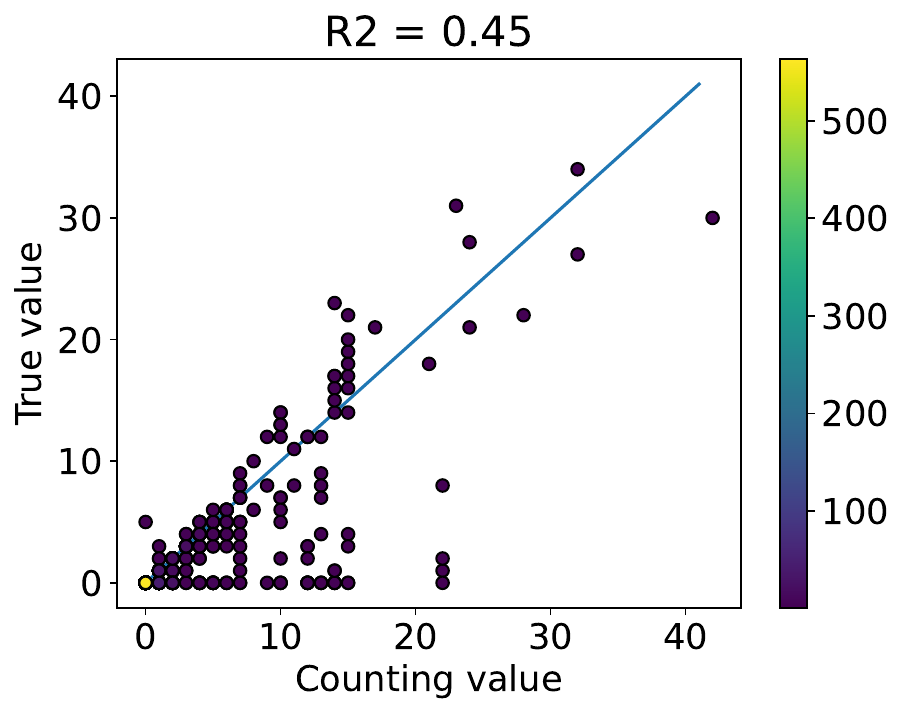}
    		\caption{\qwenvl+ G-DINO (FT-FULL)}
    \end{subfigure}
    \begin{subfigure}{0.30\linewidth}
    		\includegraphics[width=\linewidth]{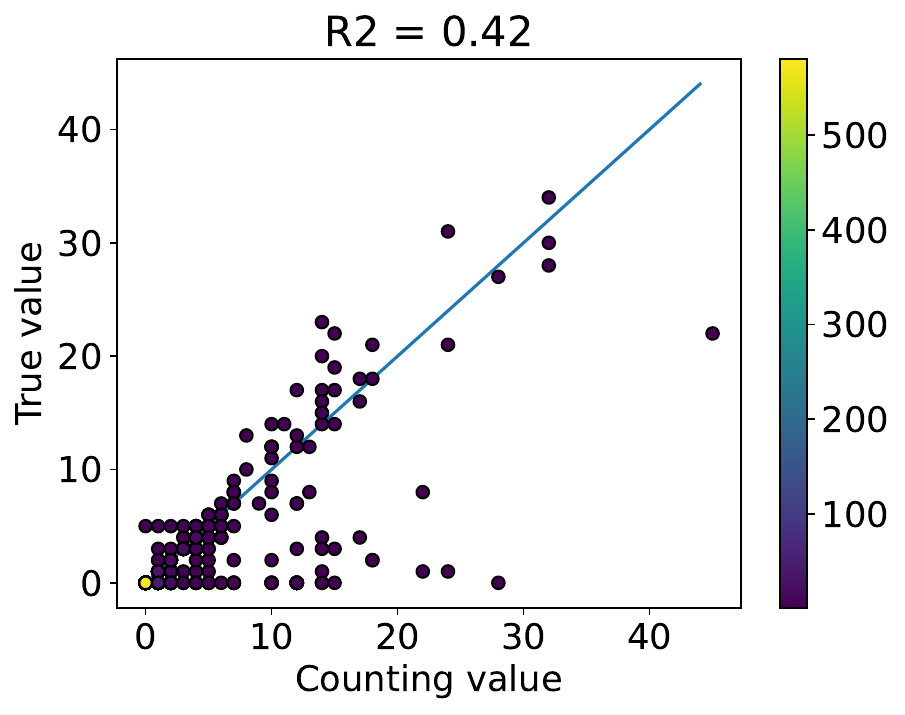}
    		\caption{\qwenvl+True bbox}
    \end{subfigure}
    \caption{Scatter plots of actual counts and counts from models on the FloodNet testing set. Blue lines are used to highlight points with correct counts. To address the overlap of numerous points, color gradients are applied to represent density, providing a clearer visual distinction.}
    \label{fig:scatter_plot_qwenvl_}
\end{figure*}

\begin{figure*}[htpb]
	\centering
    \begin{subfigure}{0.30\linewidth}
		\includegraphics[width=\linewidth]{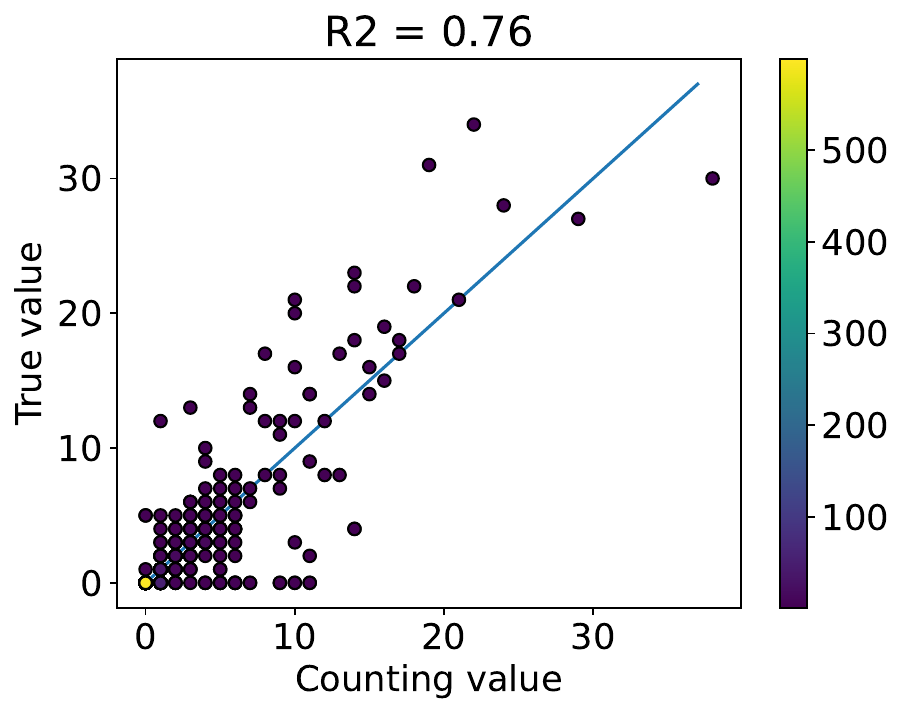}
		\caption{\internvl only}
    \end{subfigure} 
    \begin{subfigure}{0.30\linewidth}
		\includegraphics[width=\linewidth]{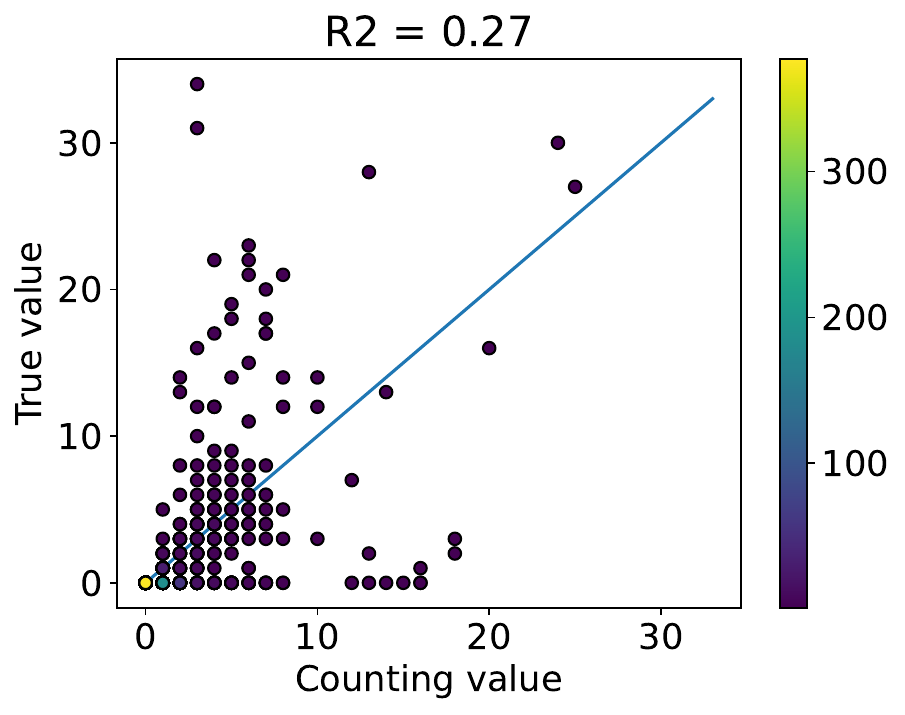}
		\caption{\internvl+ G-DINO (Not FT)}
    \end{subfigure}\\
    \begin{subfigure}{0.30\linewidth}
    		\includegraphics[width=\linewidth]{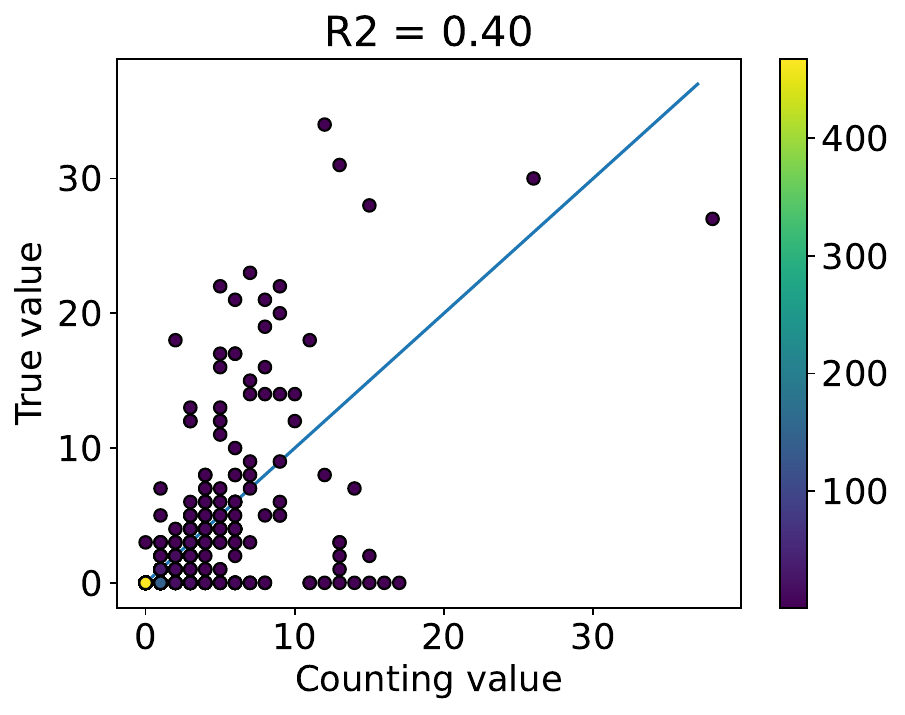}
    		\caption{\internvl+ G-DINO (FT-100)}
    \end{subfigure}
    \begin{subfigure}{0.30\linewidth}
    		\includegraphics[width=\linewidth]{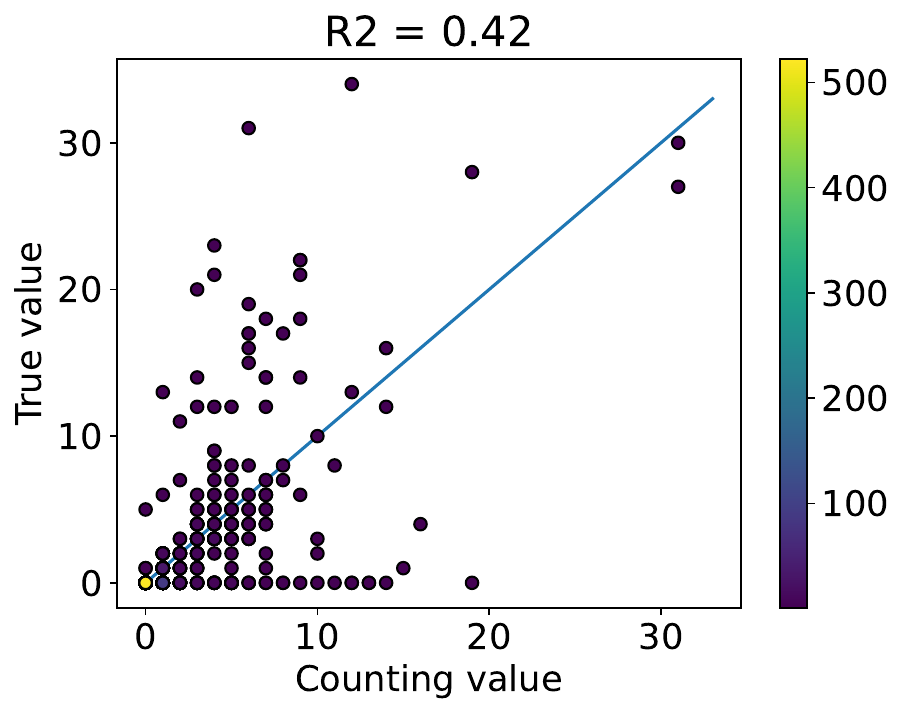}
    		\caption{\internvl+ G-DINO (FT-FULL)}
    \end{subfigure}
    \begin{subfigure}{0.30\linewidth}
    		\includegraphics[width=\linewidth]{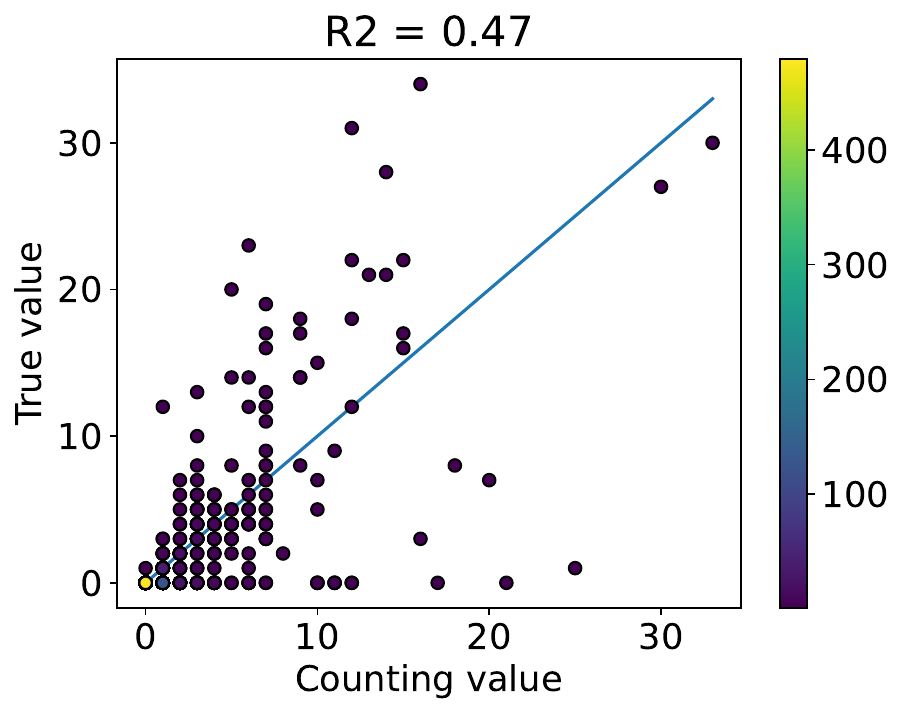}
    		\caption{\internvl+True bbox}
    \end{subfigure}
    \caption{Scatter plots of actual counts and counts from models on the FloodNet testing set.}
    \label{fig:scatter_plot_internvl_}
\end{figure*}

\begin{figure*}[htpb]
	\centering
    \begin{subfigure}{0.30\linewidth}
		\includegraphics[width=\linewidth]{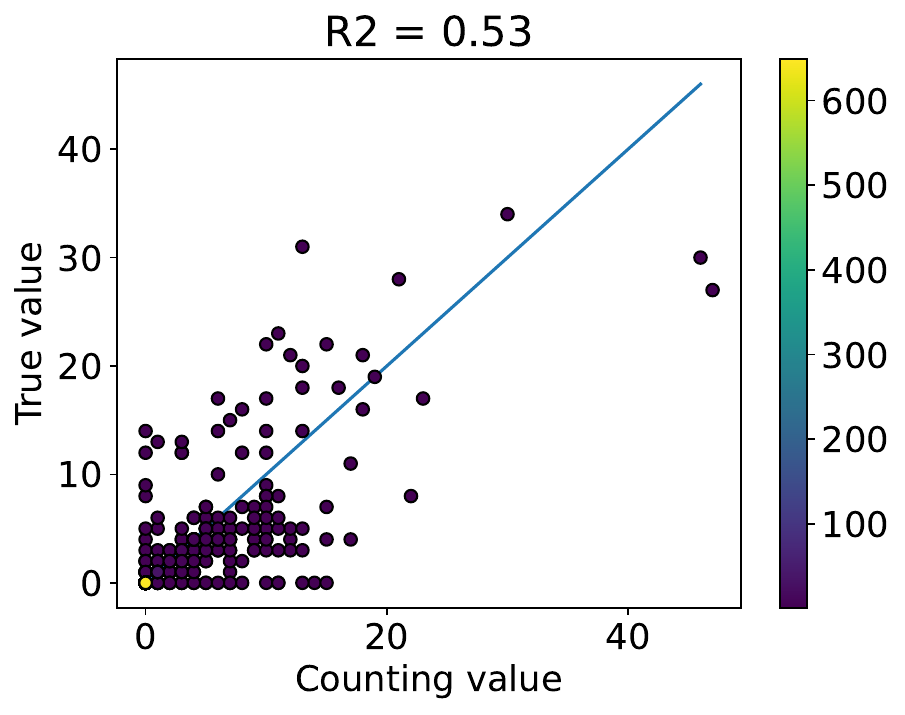}
		\caption{\gpt only}
    \end{subfigure} 
    \begin{subfigure}{0.30\linewidth}
		\includegraphics[width=\linewidth]{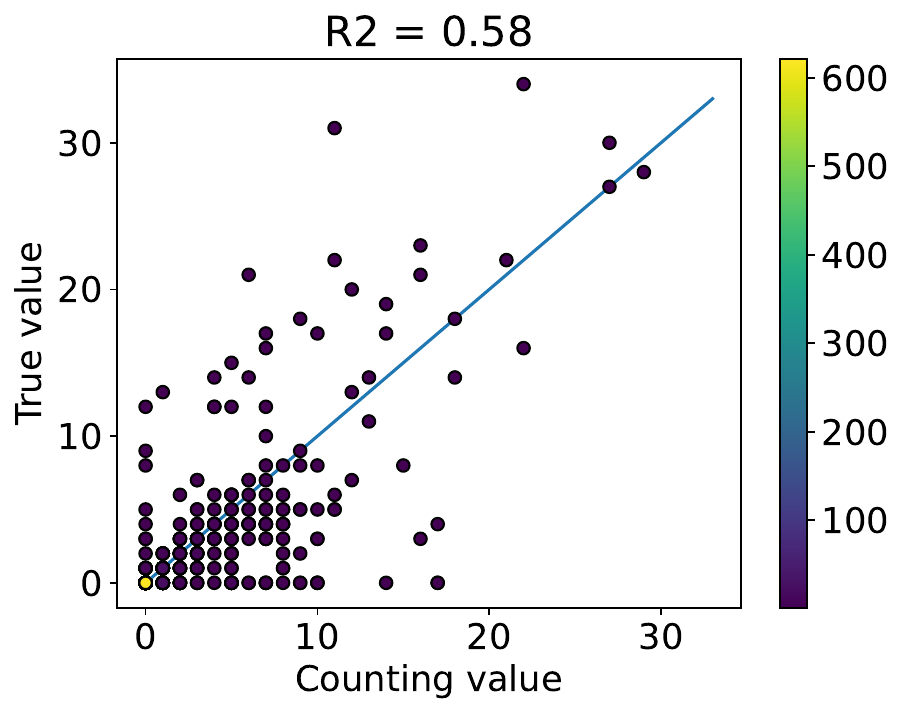}
		\caption{\gpt+ G-DINO (Not FT)}
    \end{subfigure}\\
    \begin{subfigure}{0.30\linewidth}
    		\includegraphics[width=\linewidth]{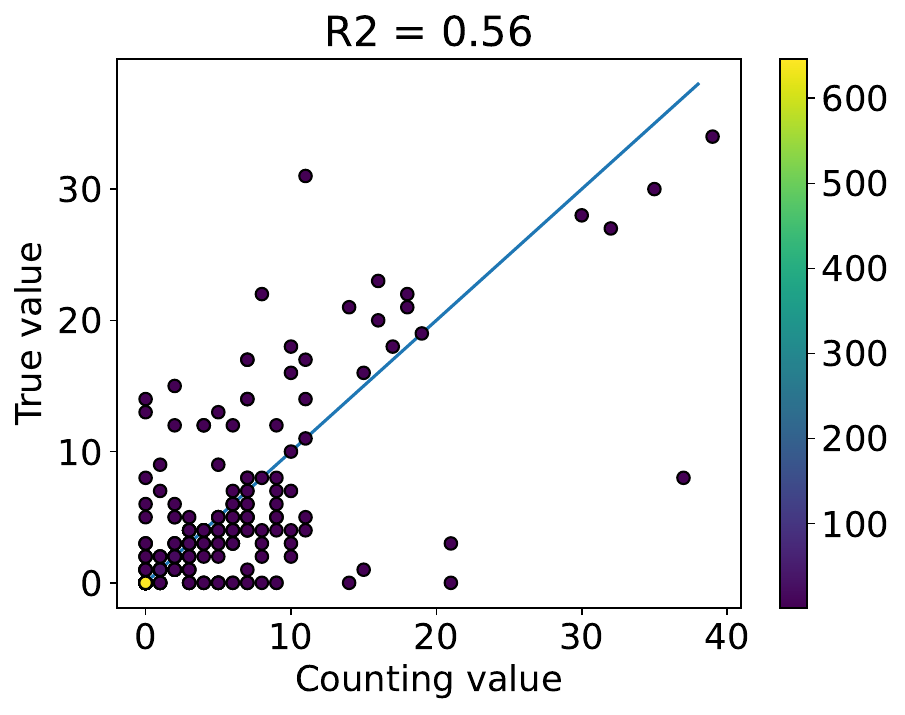}
    		\caption{\gpt+ G-DINO (FT-100)}
    \end{subfigure}
    \begin{subfigure}{0.30\linewidth}
    		\includegraphics[width=\linewidth]{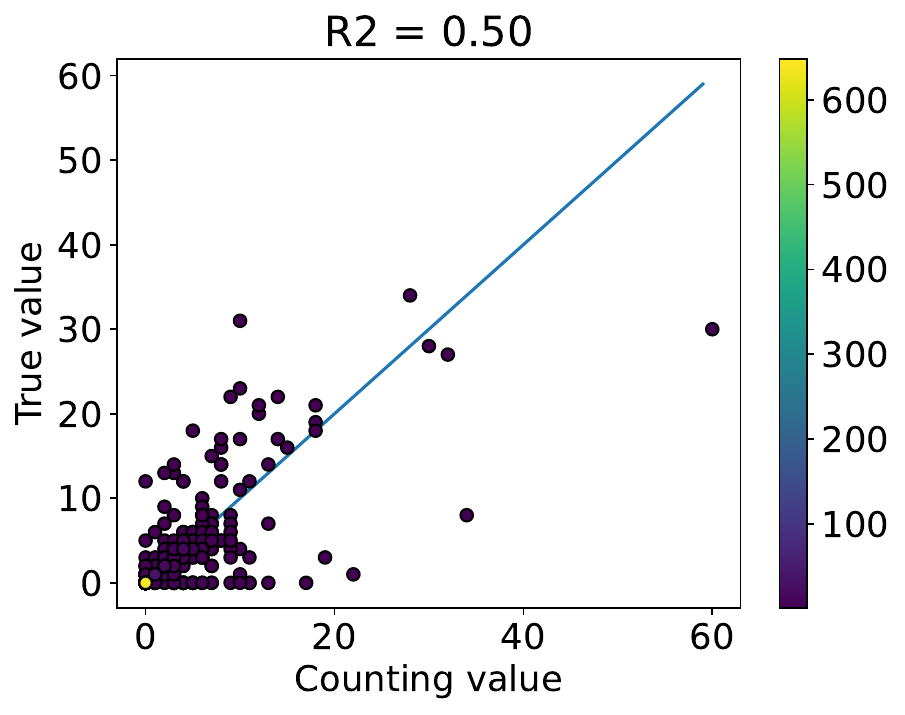}
    		\caption{\gpt+ G-DINO (FT-FULL)}
    \end{subfigure}
    \begin{subfigure}{0.30\linewidth}
    		\includegraphics[width=\linewidth]{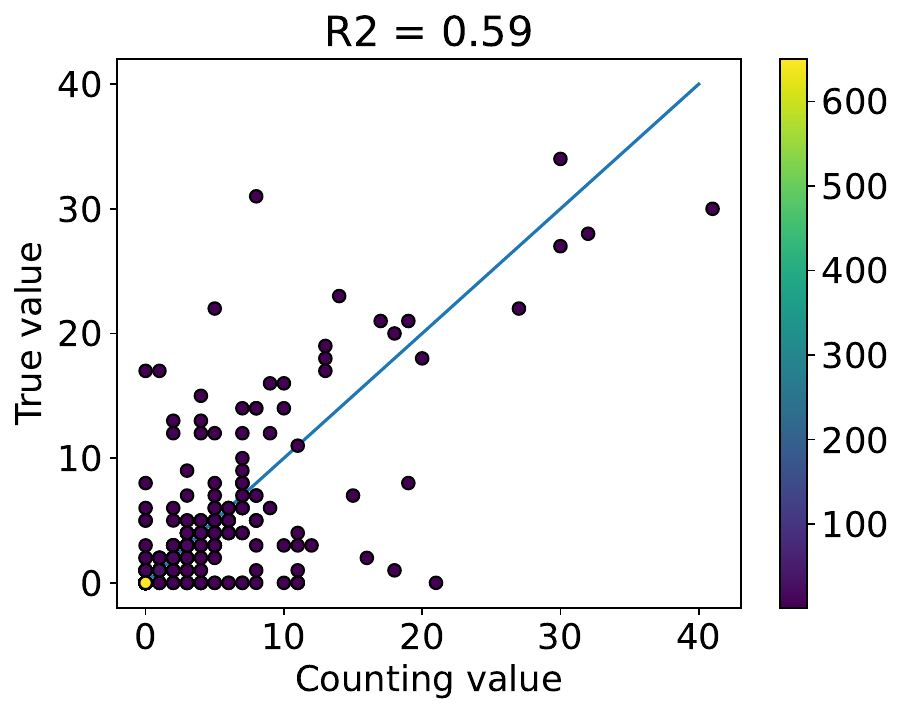}
    		\caption{\gpt+True bbox}
    \end{subfigure}
    \caption{Scatter plots of actual counts and counts from models on the FloodNet testing set. Blue lines are used to highlight points with correct counts. To address the overlap of numerous points, color gradients are applied to represent density, providing a clearer visual distinction.}
    \label{fig:scatter_plot_gpt_}
\end{figure*}

\begin{figure*}[htpb]
	\centering
    \begin{subfigure}{0.30\linewidth}
		\includegraphics[width=\linewidth]{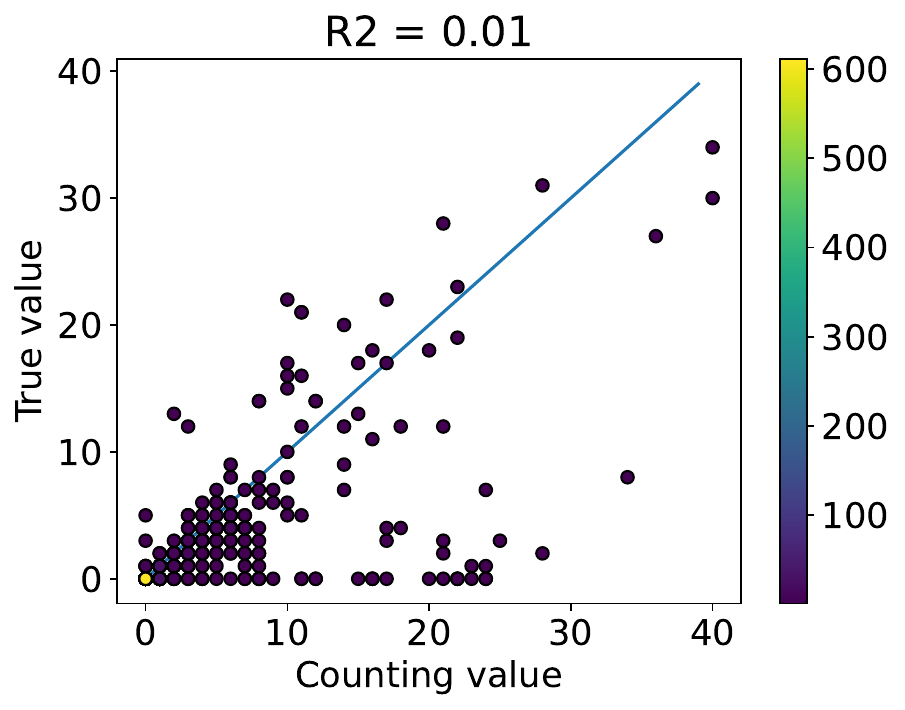}
		\caption{\gemini only}
    \end{subfigure} 
    \begin{subfigure}{0.30\linewidth}
		\includegraphics[width=\linewidth]{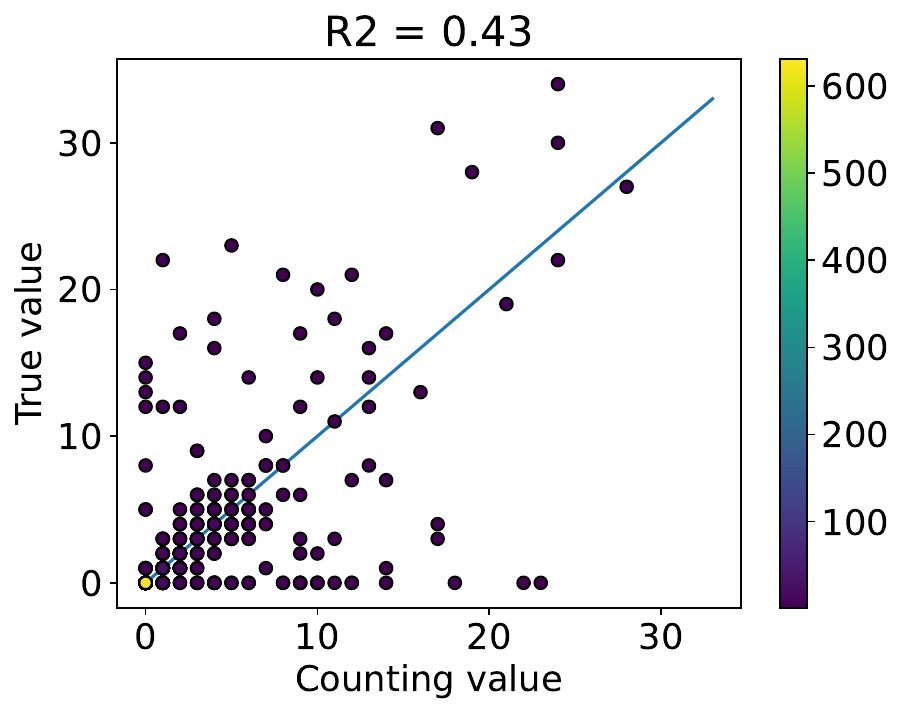}
		\caption{\gemini+ G-DINO (Not FT)}
    \end{subfigure}\\
    \begin{subfigure}{0.30\linewidth}
    		\includegraphics[width=\linewidth]{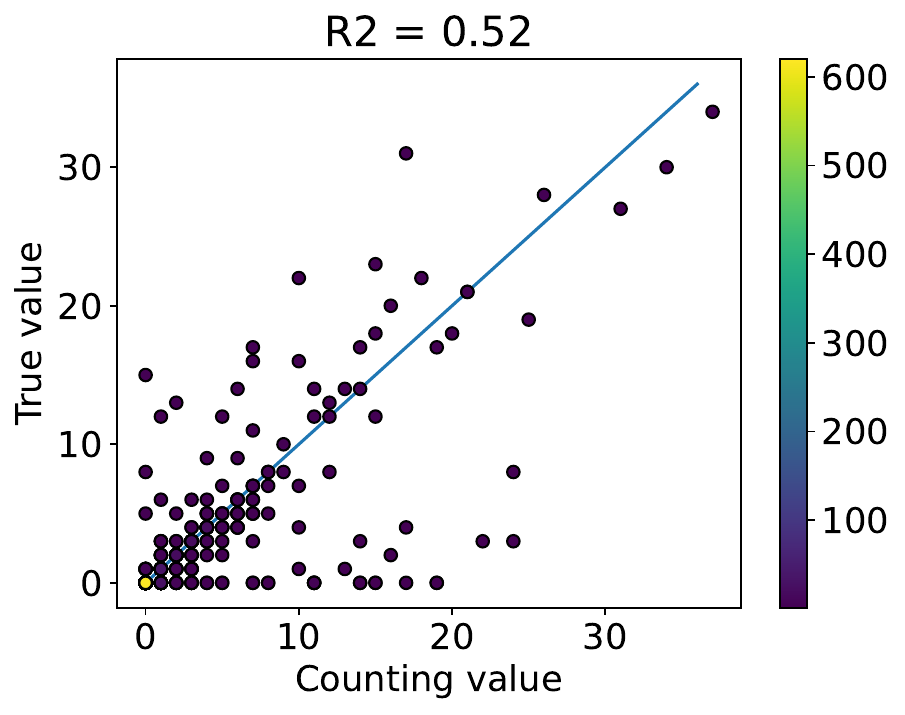}
    		\caption{\gemini+ G-DINO (FT-100)}
    \end{subfigure}
    \begin{subfigure}{0.30\linewidth}
    		\includegraphics[width=\linewidth]{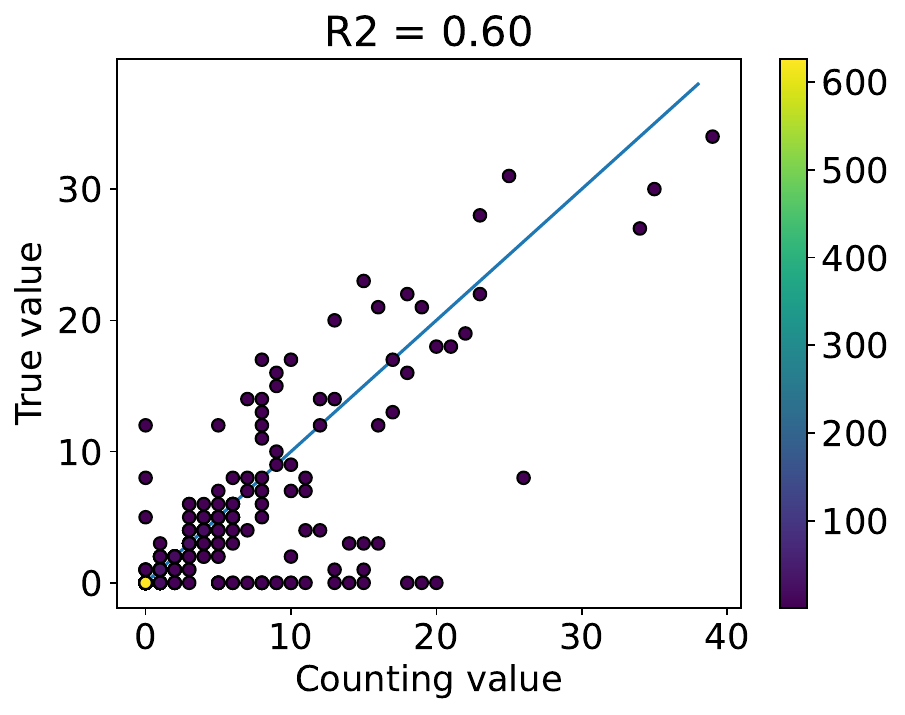}
    		\caption{\gemini+ G-DINO (FT-FULL)}
    \end{subfigure}
    \begin{subfigure}{0.30\linewidth}
    		\includegraphics[width=\linewidth]{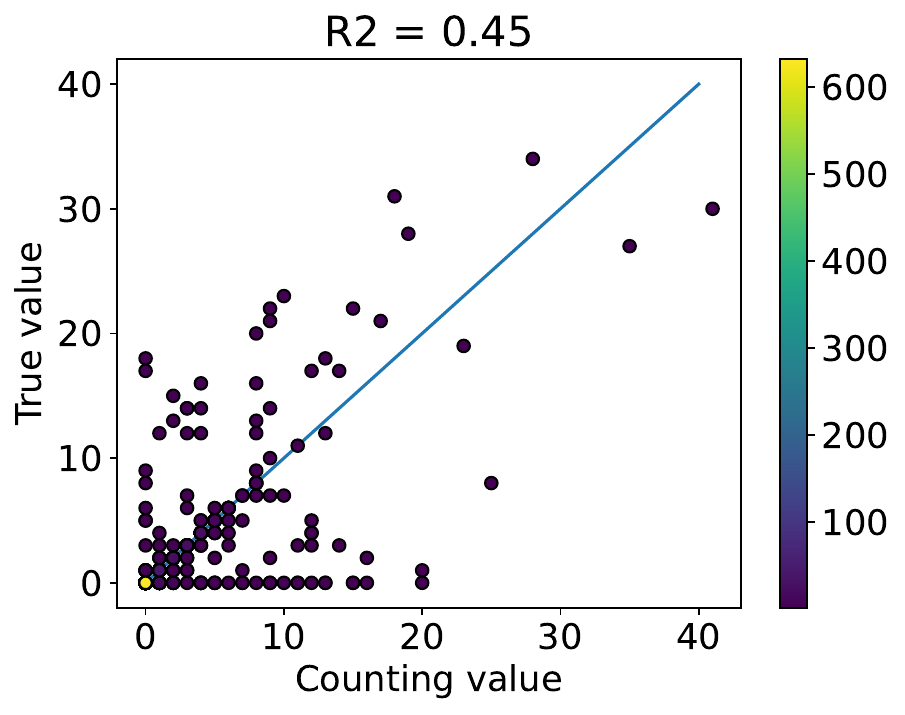}
    		\caption{\gemini+True bbox}
    \end{subfigure}
    \caption{Scatter plots of actual counts and counts from models on the FloodNet testing set.}
    \label{fig:scatter_plot_gemini_}
\end{figure*}

\begin{figure*}[htpb]
	\centering
    \begin{subfigure}{0.30\linewidth}
		\includegraphics[width=\linewidth]{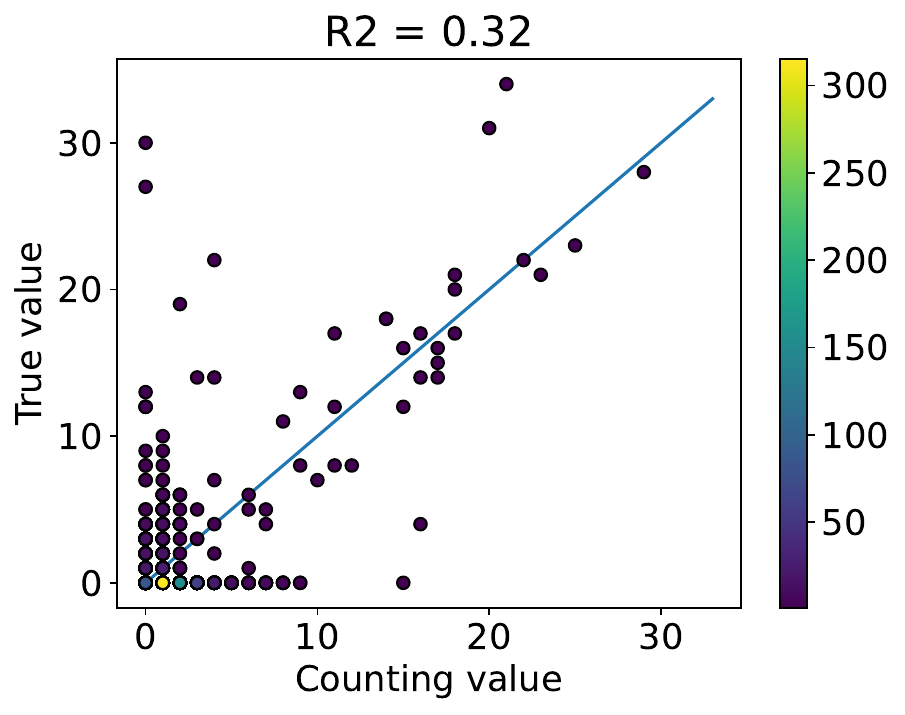}
		\caption{G-DINO* with Not FT}
    \end{subfigure}
    \begin{subfigure}{0.30\linewidth}
    		\includegraphics[width=\linewidth]{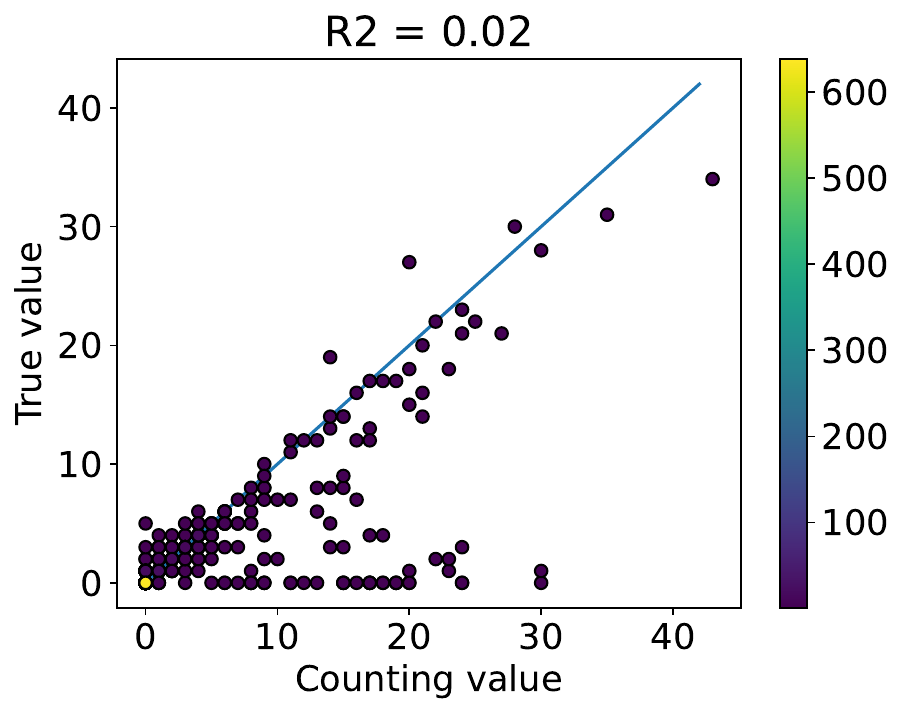}
    		\caption{G-DINO* with FT-100}
    \end{subfigure}
    \begin{subfigure}{0.30\linewidth}
    		\includegraphics[width=\linewidth]{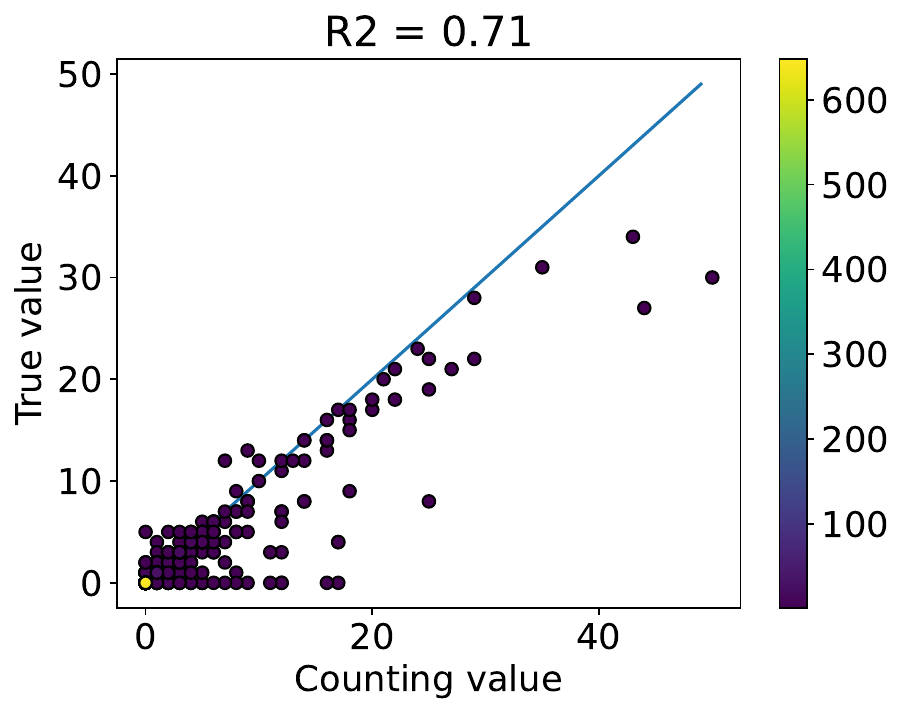}
    		\caption{G-DINO* with FT-FULL}
    \end{subfigure}
    \caption{Scatter plots of actual counts and counts from models on the FloodNet testing set.}
    \label{fig:scatter_plot_gdino_}
\end{figure*}

Figures~\ref{fig:scatter_plot_qwenvl},~\ref{fig:scatter_plot_internvl},~\ref{fig:scatter_plot_gpt},~\ref{fig:scatter_plot_gemini}, and~\ref{fig:scatter_plot_gdino}, present scatter plots of true versus predicted building counts obtained with our framework and isolated baselines on the \rNet~test set across the considered \acp{lvlm}. 
Figures~\ref{fig:scatter_plot_qwenvl_},~\ref{fig:scatter_plot_internvl_},~\ref{fig:scatter_plot_gpt_},~\ref{fig:scatter_plot_gemini_}, and~\ref{fig:scatter_plot_gdino_} present the corresponding results on the \fNet~test set.

\newpage 

\subsection{Prompt instruction}
This section presents the prompt instructions used to query the \ac{lvlm} for counting buildings in each damage category: intact, damaged, and completely destroyed for \rNet; and non-flooded and flooded for \fNet.

\begin{tcolorbox}[title=Prompt for Counting Intact Buildings (with bounding boxes)]
\small
You are given an image of a disaster-affected area with bounding boxes of general buildings detected by a computer vision model.

\textbf{Note:} The bounding boxes may be inaccurate (some buildings may be incorrectly detected or other objects may be mistakenly identified as buildings).

\textbf{Task:}
\begin{itemize}
    \item Count the number of intact buildings.
    \item Respond with the count only.
\end{itemize}

\textbf{Definition of intact buildings:}
\begin{itemize}
    \item \textbf{Intact:} buildings with no visible damage
\end{itemize}

\textbf{Guidelines:}
\begin{itemize}
    \item Count each building at most once.
    \item Include buildings that are clearly no visible damage even if they are not perfectly enclosed by a bounding box.
    \item Exclude buildings that appear damaged or completely collapsed.
\end{itemize}
\end{tcolorbox}

\begin{tcolorbox}[title=Prompt for Counting Intact Buildings (without bounding boxes)]
\small
You are given an image of a disaster-affected area.

\textbf{Task:}
\begin{itemize}
    \item Count the number of intact buildings.
    \item Respond with the count only.
\end{itemize}

\textbf{Definition of intact buildings:}
\begin{itemize}
    \item \textbf{Intact:} buildings with no visible damage
\end{itemize}

\textbf{Guidelines:}
\begin{itemize}
    \item Count each building at most once.
    \item Exclude buildings that appear damaged or completely collapsed.
\end{itemize}
\end{tcolorbox}

\begin{tcolorbox}[title=Prompt for Counting Buildings Damaged (with bounding boxes)]
\small
You are given an image of a disaster-affected area with bounding boxes of general buildings detected by a computer vision model.

\textbf{Note:} The bounding boxes may be inaccurate (some buildings may be incorrectly detected or other objects may be mistakenly identified as buildings).

\textbf{Task:}
\begin{itemize}
    \item Count the number of buildings that are damaged.
    \item Respond with the count only.
\end{itemize}

\textbf{Definition of damage types:}
\begin{itemize}
    \item \textbf{Damaged:} Buildings with damage ranging from minimal cosmetic damage to structural damage requiring extensive repairs but not completely collapsed.
\end{itemize}

\textbf{Guidelines:}
\begin{itemize}
    \item Count each building at most once.
    \item Include buildings that are clearly damaged even if they are not perfectly enclosed by a bounding box.
    \item Exclude buildings that appear intact or completely collapsed.
\end{itemize}
\end{tcolorbox}

\begin{tcolorbox}[title=Prompt for Counting Buildings Damaged (without bounding boxes)]
You are given an image of a disaster-affected area.

\textbf{Task:}
\begin{itemize}
    \item Count the number of buildings that are damaged.
    \item Respond with the count only.
\end{itemize}

\textbf{Definition of buildings damaged:}
\begin{itemize}
    \item \textbf{Damaged:} buildings with damage ranging from minimal cosmetic damage to structural damage requiring extensive repairs but not completely collapsed
\end{itemize}

\textbf{Guidelines:}
\begin{itemize}
    \item Count each building at most once.
    \item Exclude buildings that appear damaged or completely collapsed.
\end{itemize}
\end{tcolorbox}

\begin{tcolorbox}[title=Prompt for Counting Buildings Destroyed (with bounding boxes)]
\small
You are given an image of a disaster-affected area with bounding boxes of general buildings detected by a computer vision model.

\textbf{Note:} The bounding boxes may be inaccurate (some buildings may be incorrectly detected or other objects may be mistakenly identified as buildings).

\textbf{Task:}
\begin{itemize}
    \item Count the number of buildings that are completely destroyed.
    \item Respond with the count only.
\end{itemize}

\textbf{Definition of damage types:}
\begin{itemize}
    \item \textbf{Destroyed:} buildings are a total loss—completely collapsed and irreparable.
\end{itemize}

\textbf{Guidelines:}
\begin{itemize}
    \item Count each building at most once.
    \item Include buildings that are completely collapsed even if they are not perfectly enclosed by a bounding box.
    \item Exclude buildings that appear intact or damaged but not completely collapsed.
\end{itemize}
\end{tcolorbox}

\begin{tcolorbox}[title=Prompt for Counting Buildings Destroyed (without bounding boxes)]
You are given an image of a disaster-affected area.

\textbf{Task:}
\begin{itemize}
    \item Count the number of buildings that are completely destroyed.
    \item Respond with the count only.
\end{itemize}

\textbf{Definition of buildings destroyed:}
\begin{itemize}
    \item \textbf{Destroyed:} buildings are a total loss—completely collapsed and irreparable.
\end{itemize}

\textbf{Guidelines:}
\begin{itemize}
    \item Count each building at most once.
    \item Exclude buildings that appear intact or damaged but not completely collapsed.
\end{itemize}
\end{tcolorbox}

\begin{tcolorbox}[title=Prompt for Counting Buildings Flooded (with bounding boxes)]
\small
You are given an image of a disaster-affected area with bounding boxes of general buildings detected by a computer vision model.

\textbf{Note:} The bounding boxes may be inaccurate (some buildings may be incorrectly detected or other objects may be mistakenly identified as buildings).

\textbf{Task:}
\begin{itemize}
    \item Count the number of buildings that are flooded.
    \item Respond with the count only.
\end{itemize}

\textbf{Definition of buildings flooded:}
\begin{itemize}
    \item A building is classified as flooded when at least one side of the building is touching the flood water.
\end{itemize}

\textbf{Guidelines:}
\begin{itemize}
    \item Count each building at most once.
    \item Include buildings that are clearly flooded even if they are not perfectly enclosed by a bounding box.
    \item Exclude buildings that show no visible signs of flooding.
\end{itemize}
\end{tcolorbox}

\begin{tcolorbox}[title=Prompt for Counting Buildings Flooded (without bounding boxes)]
You are given an image of a disaster-affected area.

\textbf{Task:}
\begin{itemize}
    \item Count the number of buildings that are flooded.
    \item Respond with the count only.
\end{itemize}

\textbf{Definition of buildings flooded:}
\begin{itemize}
    \item A building is classified as flooded when at least one side of the building is touching the flood water.
\end{itemize}

\textbf{Guidelines:}
\begin{itemize}
    \item Count each building at most once.
    \item Exclude buildings that show no visible signs of flooding.
\end{itemize}
\end{tcolorbox}

\begin{tcolorbox}[title=Prompt for Counting Buildings Non-Flooded (with bounding boxes)]
\small
You are given an image of a disaster-affected area with bounding boxes of general buildings detected by a computer vision model.

\textbf{Note:} The bounding boxes may be inaccurate (some buildings may be incorrectly detected or other objects may be mistakenly identified as buildings).

\textbf{Task:}
\begin{itemize}
    \item Count the number of buildings that are non-flooded.
    \item Respond with the count only.
\end{itemize}

\textbf{Definition of buildings non-flooded:}
\begin{itemize}
    \item A building is classified as non-flooded when no side of the building is touching the flood water.
\end{itemize}

\textbf{Guidelines:}
\begin{itemize}
    \item Count each building at most once.
    \item Include buildings that are clearly non-flooded even if they are not perfectly enclosed by a bounding box.
    \item Exclude buildings that show visible signs of flooding.
\end{itemize}
\end{tcolorbox}

\begin{tcolorbox}[title=Prompt for Counting Buildings Non-Flooded (without bounding boxes)]
You are given an image of a disaster-affected area.

\textbf{Task:}
\begin{itemize}
    \item Count the number of buildings that are flooded.
    \item Respond with the count only.
\end{itemize}

\textbf{Definition of building non-flooded:}
\begin{itemize}
    \item A building is classified as non-flooded when no side of the building is touching the flood water.
\end{itemize}

\textbf{Guidelines:}
\begin{itemize}
    \item Count each building at most once.
    \item Exclude buildings that show visible signs of flooding.
\end{itemize}
\end{tcolorbox}

\end{document}